\documentclass[final,3p,times]{elsarticle}
\usepackage{natbib} % 
\usepackage{hyperref}
\usepackage{algorithm}
\usepackage{algpseudocode}
\usepackage{algorithmicx}
\usepackage{subcaption}
\usepackage{adjustbox}   % 可选，如果要自动缩放
\usepackage{booktabs}
\usepackage{longtable}
\usepackage{multirow}
\usepackage{amssymb}
\usepackage{amsthm}
\usepackage{amsmath}

\usepackage{booktabs}
\usepackage[normalem]{ulem}
\useunder{\uline}{\ul}{}
\journal{}
\makeatletter
\def\ps@pprintTitle{%
	\let\@oddhead\@empty
	\let\@evenhead\@empty
	\let\@oddfoot\@empty
	\let\@evenfoot\@empty}
\makeatother

\begin{document}

\begin{frontmatter}



\title{Dual Refinement Cycle Learning: Unsupervised Text Classification of Mamba and Community Detection on Text Attributed Graph} %% Article title


\author[a]{Hong Wang} %% Author name
\author[a]{Yinglong Zhang\corref{*}}
\ead{zhang_yinglong@126.com}
\cortext[*]{Corresponding author}
\author[a]{Hanhan Guo}
\author[a]{Xuewen Xia}
\author[a]{Xing Xu}

\affiliation[a]{
	organization={College of Physics and Information Engineering, Minnan Normal University},
	city={Zhangzhou},
	state={Fujian},
	postcode={363000},
	country={China}}

%% Abstract
\begin{abstract}
Pretrained language models offer strong text understanding capabilities but remain difficult to deploy in real-world text-attributed networks due to their heavy dependence on labeled data. Meanwhile, community detection methods typically ignore textual semantics, limiting their usefulness in downstream applications such as content organization, recommendation, and risk monitoring. To overcome these limitations, we present Dual Refinement Cycle Learning (DRCL), a fully unsupervised framework designed for practical scenarios where no labels or category definitions are available. DRCL integrates structural and semantic information through a warm-start initialization and a bidirectional refinement cycle between a GCN-based Community Detection Module (GCN-CDM) and a Text Semantic Modeling Module (TSMM). The two modules iteratively exchange pseudo-labels, allowing semantic cues to enhance structural clustering and structural patterns to guide text representation learning without manual supervision. Across several text-attributed graph datasets, DRCL consistently improves the structural and semantic quality of discovered communities. Moreover, a Mamba-based classifier trained solely from DRCL's community signals achieves accuracy comparable to supervised models, demonstrating its potential for deployment in large-scale systems where labeled data are scarce or costly. The code is available at \url{https://github.com/wuanghoong/DRCL.git}.

\end{abstract}



%% Keywords

\end{frontmatter}



\section{Introduction}
Despite the ubiquity of pretrained language models, practical text classification relies heavily on supervised fine-tuning with labeled data. Such models assume that the category set is known and fixed in advance, and that a substantial amount of high-quality annotated samples is available to adapt and calibrate the model \cite{RN6}. However, this assumption is often difficult to satisfy in dynamic real-world scenarios. Obtaining high-quality annotations is cost-prohibitive, and categories often evolve continuously \cite{RN8}. Consequently, a critical challenge arises: enabling a language model to perform reliable text classification in the complete absence of ground-truth labels and prior knowledge regarding category quantity. 

In graph–text coupled complex networks, both topological structure and textual content are present simultaneously \cite{RN7}. For example, in academic citation networks, each node is associated with the title and abstract of a paper, and the edges depict citation relations. In social networks, user nodes contain profile descriptions and interest tags, and the edges reflect following or interaction behaviors. In knowledge graphs, entities and relations are often accompanied by descriptive text. In such real-world networks, annotations are often scarce, and it is difficult to precisely specify in advance a reasonable set of categories and its size. However, these categories are implicitly embedded in the grouping patterns that emerge from nodes which are semantically similar and strongly connected in the graph structure. How to enable a language model to learn to effectively distinguish texts using only the network structure and the associated text, when both the labels and the number of categories are unknown, constitutes the core challenge addressed in this paper. 

Community detection in complex networks is a key graph mining technique used to uncover latent grouping structures. Its objective is to identify subsets of nodes that are densely connected internally and relatively sparse externally \cite{RN3}. Early approaches mainly relied on topological structure, such as spectral partitioning and modularity maximization \cite{RN2}. More recent deep attributed graph clustering methods model both network structure and node attribute features simultaneously \cite{RN4}. These methods assume that the true number of communities is known in advance and characterize a fixed set of latent groups through joint modeling of structure and attributes. While existing community detection methods have advanced by jointly modeling structure and attributes, they face two primary limitations. First, they predominantly focus on topology and numerical features, thereby underutilizing the rich fine-grained semantics embedded in node texts \cite{RN5}. Second, current integrations of language models into graph tasks typically necessitate ground-truth labels or a known number of clusters, limiting their applicability in open environments where such priors are absent. 

To bridge this gap, we propose a unified perspective that explicitly couples unsupervised text classification with graph community detection. We hypothesize that even in the absence of ground-truth labels and prior knowledge of category quantity, the category discovery capability of a deep community detection model can provide essential supervisory signals to guide the language model. Conversely, the semantic parsing ability of the language model can be leveraged to enhance the discriminability of community detection. Through this reciprocal interaction, community partitioning is no longer reliant solely on structural and attribute features but is instead jointly optimized by structural and semantic perspectives, yielding more consistent grouping results. 

Guided by this hypothesis, we introduce the Dual Refinement Cycle Learning (DRCL) framework, which constructs a bidirectional closed-loop optimization mechanism. Under the constraints of proto-community signals derived from the graph structure, the framework utilizes an unsupervised GCN-based community detection module to adaptively generate community pseudo-labels. These labels serve as supervisory signals for the language model, enabling it to acquire semantically discriminative text representations and pseudo-labels without manual annotation. In turn, the learned semantic information is fed back into the community detection module, updating its feature inputs and proto-community signals. This feedback provides discriminative semantic cues for community partitioning and thereby improves the accuracy and consistency of the resulting community assignments. Through repeated iterative cycles, the DRCL framework progressively enhances both the language model and the community detection model at the structural and semantic levels, thereby achieving a coordinated and mutually reinforcing optimization of the two. 

The main contributions of this paper are summarized as follows:
\begin{enumerate}
	\item \textbf{A novel Dual Refinement Cycle Learning framework is introduced.} A bidirectional closed-loop optimization mechanism is constructed by integrating a GCN-based Community Detection Module (GCN-CDM) with a Text Semantic Modeling Module (TSMM). This design enables continuous complementary learning between topological and semantic information, allowing the two modules to be jointly trained and iteratively refined.
	\item \textbf{Effective training of the Mamba language model is achieved under unlabeled and category-unknown conditions.} DRCL uses high-quality community detection results as supervisory signals to drive the training of the Mamba model. This design eliminates the need for predefined category numbers and manual annotations, enabling the model to complete text classification without relying on human-labeled data.
	\item \textbf{A unified unsupervised learning mechanism for graph neural networks and language models is established.} Within DRCL, the GCN and the language model form a closed-loop learning process in which the output of one module serves as the input to the other. This reciprocal refinement unifies topological connectivity and deep semantic information, significantly enhancing the structural interpretability and semantic discriminability of community detection.
	\item \textbf{Extensive experiments on multiple text-attributed graph datasets demonstrate the effectiveness of the proposed approach.} The method achieves stable and substantial improvements in both community detection quality and text classification performance. Incorporating textual information markedly enhances the semantic discriminability of community assignments, while community-guided training enables the Mamba model to reach or even surpass the performance of supervised training with medium-scale human-labeled data. This demonstrates the practical value of the framework as well as its strong potential for generalization.
\end{enumerate}

\section{Related Work}
\subsection{Community Detection with Deep Learning}
Research on community detection based on deep learning has undergone a progression from early exploratory attempts to deeper integration and paradigm innovation. In the exploratory stage, deep models were first used to learn low-dimensional node embeddings, and traditional clustering algorithms were then applied to the learned embeddings. Kipf and Welling introduced VGAE \cite{RN9}, which stands as a seminal work in this research field. The model employs a graph convolutional encoder together with an inner-product decoder, enabling end-to-end unsupervised learning of both graph structure and node attributes for the first time. The resulting embeddings have been widely adopted in subsequent clustering studies, and the method quickly established itself as a benchmark in the field. Inspired by VGAE, subsequent research shifts its focus toward improving the robustness and discriminability of graph embeddings, which in turn gives rise to a variety of enhanced paradigms. For example, the Adversarially Regularized Graph Autoencoder (ARGA) \cite{RN10} introduces a generative adversarial framework that drives the latent codes to match a prior distribution, thereby learning embeddings with stronger generalization ability. The Symmetric Graph Convolutional Autoencoder (SGAE) \cite{RN11} constructs a symmetric encoder–decoder architecture to strengthen the model’s capability in reconstructing and preserving graph structural information. However, these methods face an inherent limitation: their representation learning objectives fundamentally diverge from the ultimate goal of community detection, which often results in embeddings that are suboptimal for capturing community structures. 

To overcome these limitations, research has shifted toward designing end-to-end models that place community structure at the core. Modularity, as the most classic measure of community quality, has therefore become the central optimization objective in many of these models. DGCluster \cite{RN12} makes the modularity maximization objective differentiable and combines it with GNN-based soft community similarity, enabling efficient deep clustering without requiring a predefined number of clusters. Similarly, the modularity-aware graph autoencoder \cite{RN13} incorporates modularity constraints directly into both its message-passing mechanism and loss function, thereby integrating community information more deeply into the encoding process. As research progresses, the role and applications of modularity have continued to evolve and deepen. MAGI \cite{RN14} analyzes modularity from a contrastive learning perspective and shows that modularity maximization corresponds to a specific community-aware contrastive learning task. In this way, the model can capture higher-order proximity without relying on graph augmentations. In addition, MOMCD \cite{RN15} introduces motif weights to extend modularity from edge-level structures to higher-order structures, enabling more accurate detection of functional units in networks. 

Meanwhile, other learning paradigms and technical routes have greatly enriched the landscape of deep community detection. The introduction of contrastive learning has significantly enhanced the discriminative ability of models. CommDGI \cite{RN16} incorporates both community-level and graph-level mutual information objectives on top of a GNN encoder. DCLN \cite{RN17} designs a two-level contrastive mechanism at the node and feature levels to mitigate representation collapse. CPGCL \cite{RN18} further innovates by using the community probability distribution to dynamically guide the contrastive learning process, forming a self-supervised loop in which community assignment and representation learning reinforce each other. DCGL \cite{RN19} addresses clustering scenarios without prior graph structure by employing a pseudo-siamese network to jointly learn structural and attribute information, and by optimizing cluster structures through local–global contrastive learning. For multi-view settings, SGCMC \cite{RN20} employs a graph attention autoencoder with multi-view contrastive learning to achieve end-to-end soft clustering. To pursue extreme efficiency and scalability, SCGC \cite{RN21} adopts a lightweight MLP architecture instead of a complex GNN and performs contrastive learning without data augmentation, enabling efficient clustering on large-scale graphs. 

Existing deep community detection methods primarily rely on structural information and numerical attributes, making it difficult for them to exploit the fine-grained semantic content contained in node texts. They also typically require a predefined number of communities and lack the ability for continuous refinement. In contrast, DRCL incorporates semantic representations from a language model into its iterative process, allowing community partitioning to be constrained jointly by structural and semantic signals. Through alternating updates of pseudo-labels, DRCL can dynamically adjust community boundaries and category structures, fundamentally overcoming the static and single-modal limitations of traditional deep community detection models.

\subsection{Language Models}

The evolution of language models mirrors the broader trajectory of natural language processing toward more general intelligence, and their development clearly traces a path from architectural innovation to paradigm establishment and ecosystem expansion. This trajectory began with the introduction of the Transformer architecture \cite{RN22}. Its self-attention mechanism provides strong parallelism and effective modeling of long sequences, replacing traditional RNNs \cite{RN23} and LSTMs \cite{RN24}. This shift established the core foundation for subsequent large-scale models. After the advent of the Transformer architecture, the pre-training paradigm quickly diverged into two distinct paths. The first path is represented by BERT \cite{RN25} and its successors such as RoBERTa \cite{RN26} and ALBERT \cite{RN27}. These bidirectional encoder architectures rely on masked language modeling to produce high-quality contextual representations and achieve strong performance on understanding-oriented tasks. Specialized variants such as SciBERT \cite{RN28} further demonstrate the effectiveness of domain-adaptive pre-training. The second path is characterized by the autoregressive decoder architecture exemplified by the GPT series \cite{RN29}. Through the successive development of GPT-2, GPT-3, and GPT-4 \cite{RN30}\cite{RN31}\cite{RN32}, both the parameter scale and generative capability of these models have expanded rapidly. This progression not only verifies scaling laws but also introduces emergent in-context learning abilities that fundamentally transform the paradigm of human–computer interaction. 

As model scaling approaches its limits, recent research has begun to exhibit a diversified trend that includes architectural innovation, vertical specialization, and open-source collaboration. At the architectural level, the state-space–based Mamba model \cite{RN33} and its optimized variant Mamba-2 \cite{RN34} challenge the dominance of the Transformer with their linear computational complexity in long-sequence processing, opening new directions for efficient inference. In parallel, a strong open-source ecosystem has become a central driving force for the democratization of technology. Meta’s LLaMA series \cite{RN35}\cite{RN36}\cite{RN37} and its instruction-tuned variants such as Alpaca and Vicuna \cite{RN39} have lowered the barrier to research. In addition, Google’s Gemma \cite{RN40}, Mistral AI’s Mistral models \cite{RN41}, and a range of models developed by Chinese teams, including Qwen \cite{RN42}\cite{RN43}\cite{RN44} and ChatGLM \cite{RN45}, are engaged in intense competition and innovation in terms of performance, multilingual capability, and applications in vertical domains. Beyond these general-purpose language models, the Jina series \cite{RN46}\cite{RN47}\cite{RN48}\cite{RN49}, which focuses on developing embedding models for multimodal retrieval, marks a shift in large model research from a single text modality toward the unified intelligent agent systems that combine perception and action. This line of work aims to enable large models to invoke external tools and autonomously plan complex tasks.

Traditional language models typically rely on ground-truth annotations or a predefined number of categories for text classification, and they are not well suited to participate directly in graph-structured tasks. DRCL addresses this limitation by using the pseudo-labels produced by GCN-CDM as supervisory signals, enabling Mamba to perform effective classification learning under fully unlabeled conditions. At the same time, the semantic representations generated by the language model feed back into the framework and enhance the discriminability of community structures. This joint refinement of language and graph models is a capability that existing one-directional enhancement approaches for language models do not possess.

\subsection{Text-Attributed Graphs}

Text-attributed graphs (TAG) integrate textual semantics into graph structures and are widely used in social networks, knowledge graphs, recommendation systems, and other domains. Traditional approaches typically rely on graph neural networks to aggregate structural signals while encoding textual content only at a shallow level. With the rise of language models, research has rapidly shifted toward LLM-for-TAG methods, which leverage large language models to extract deeper textual semantics and thereby improve performance on graph-related tasks \cite{RN50}. TAPE \cite{RN51} generates textual explanations for nodes using an LLM and incorporates them as enhanced features into downstream GNNs for node classification, significantly enriching the semantic expressiveness of graph representations. Locle \cite{RN7} uses GNNs to filter nodes and employs an LLM to produce pseudo-labels. It further introduces a two-step error-correction paradigm that combines graph rewiring with self-training, and refines noisy edges through Dirichlet energy minimization to progressively purify labels during iterative training.

In terms of model architecture, research has gradually shifted from simple feature augmentation to deeper forms of joint modeling. Structure-aware prompting methods \cite{RN52}\cite{RN53} embed graph structural information directly into prompt templates, enabling large language models to perceive graph topology during text generation or reasoning and thereby improving their performance on graph-related tasks. Furthermore, structure-aware language models such as GIANT \cite{RN54} and Edgeformers \cite{RN55} incorporate graph structure directly into the Transformer’s encoding process, achieving unified modeling of textual content and graph topology.

To improve the generalization ability and cross-task adaptability of TAG models, researchers have begun to explore the construction of graph foundation models \cite{RN56}. OFA \cite{RN57} proposes a unified formulation of graph tasks and uses a language model to describe nodes and edges in natural language, enabling zero-shot graph learning without task-specific fine-tuning. Methods such as GraphCLIP \cite{RN58} and UniGLM \cite{RN59} leverage contrastive learning and perform pre-training on multi-domain TAG datasets to acquire transferable joint graph–text representations. In addition, MARK \cite{RN60} introduces a multi-agent collaboration mechanism in which several LLMs act as concept, generation, and reasoning agents to support cluster-level semantic induction and text alignment. 

Meanwhile, another line of research focuses on enhancing the reasoning capabilities of language models by leveraging graph structure. Treating TAGs as an external source of knowledge, researchers have proposed several integration mechanisms, such as embedding-based fusion \cite{RN61}, soft prompt–based fusion \cite{RN62}, and retrieval-augmented generation \cite{RN63}\cite{RN64}. These approaches help mitigate hallucination in knowledge-intensive tasks and improve LLM performance on multi-hop reasoning, question answering, and decision-making tasks.

Most existing TAG methods adopt a one-way pipeline that enhances either semantics or structure, but they cannot achieve joint optimization of the two. DRCL establishes a bidirectional cycle process in which structural pseudo-labels guide semantic learning and semantic representations are fed back to refine structural patterns. The process allows graph structure and textual semantics to improve together over multiple iterations. Compared with static fusion or single-round self-training approaches, the bidirectional refinement mechanism of DRCL provides more stable gains in both the semantic consistency and structural soundness of community partitioning.

\section{Preliminaries}
\noindent \textbf{Definition 1.} \textbf{Text-Attributed Graphs.} A text-attributed graph is represented as a quadruple \(G=(V, E, X, T)\). Here \(V=\{v_1, v_2, ..., v_n\}\) is the node set and \(|V|=n\) is the number of nodes. The edge set \(E \subseteq V \times V \) describes pairwise relations: an edge \(e_{ij} = (v_i, v_j ) \in  E\) indicates that nodes \(v_i\) and node \(v_j\) are adjacent, and we denote the total number of edges by \(|E|=M\). The matrix \(X=\left\{x_{1}, x_{2}, x_{3}, \ldots, x_{n}\right\} \in \mathbb{R}^{n \times d}\) collects the attribute features of all nodes, where the \(i\)-th row vector \(x_i \in \mathbb{R}^d\) is the \(d\)-dimensional feature representation of node \(v_i\). The set \(T=\{t_1, t_2, ..., t_n\}\) stores the textual content attached to the nodes, and each \(t_i\) corresponds to the text associated with node \(v_i\). The combinatorial structure of \(G\) is encoded by the adjacency matrix \(A \in\{0,1\}^{n \times n}\) with entries \(a_{ij}=1\) if \((v_i, v_j) \in E\) and \(a_{ij}=0\) otherwise. From \(A\) we obtain the diagonal degree matrix \(D=\operatorname{diag}\left(d_{1}, d_{2}, d_{3}, \ldots, d_{n}\right)\), where each diagonal element \(d_{i}=\sum_{j=1}^{n} a_{i j}\) counts the number of neighbors directly connected to node \(v_i\).

\noindent \textbf{Definition 2.} \textbf{Graph Convolutional Networks.} Graph Convolutional Networks (GCNs) implement neural message passing on graphs and can be viewed as a particular form of spectral filtering of graph signals. 

The commonly used propagation rule arises from a first-order Chebyshev polynomial approximation of the spectral graph convolution, leading to an efficient neighborhood aggregation scheme in the spatial domain. The update rule of GCN is
\begin{equation}
	H^{(l+1)}=GCN(H^{(l)},A)=\delta(\tilde{A}H^{(l)}W^{(l)})
	\label{GCN}
\end{equation}
where \(H^{(l)} =\{h_1^{(l)}, h_2^{(l)}, ..., h_n^{(l)}\} \in \mathbb{R}^{n \times d'}\) denotes the node representation matrix of the \(l\)-layer, and the input of the first layer is \(H^{(0)} =X\). The adjacency matrix with self-loops is \(\hat{A}=A+I \), where \(I\) is the identity matrix, ensuring that each node also propagates its own feature during aggregation. The corresponding degree matrix \(\hat{D}\) is diagonal with entries \(\hat{D}_{ii}=  {\textstyle \sum_{j}^{}}\hat{A}_{ij}\). The normalized adjacency matrix is then defined as \(\tilde{A} = \hat{D}^{-1/2} \hat{A} \hat{D}^{-1/2}\) which stabilizes the training process and alleviates gradient explosion or vanishing. The matrix \( W^{(l)}\) is the learnable weight parameter of the \(l\)-th layer, and \(\sigma(\cdot) \) denotes a nonlinear activation function (PReLU in our implementation), enhancing the model’s expressive power and the discriminability of node representations.

\noindent \textbf{Definition 3.} \textbf{Mamba Model.} Mamba is a sequence modeling backbone built on the Selective State Space Model. It performs recursive updates of content-adaptive state-space parameters along the temporal dimension. At the same time, by combining the selective scanning algorithm, it achieves linear-time modeling of very long sequences without relying on attention mechanisms.

Let the input sequence be \(x_t \in \mathbb{R}^{d_{in}}\), the hidden state be \(h_t \in \mathbb{R}^N\), and the output be \(y_t \in \mathbb{R}^{d_{out}}\). The discrete-time recursive form of Mamba at time step \(t\) is given by:
\begin{equation}
	h_t= \bar{A} h_{t-1} + \bar{B} x_{t-1}, y_t = Ch_t
	\label{hidden}
\end{equation}
where \(\bar{A}=\mathrm{exp}(\bigtriangleup A) \), \(\bar{B} = (\bigtriangleup A)^{-1} ( \mathrm{exp}(\bigtriangleup A) -1) \cdot \bigtriangleup B\) , and \(C\) are the system matrices. \(\bigtriangleup \) is a learnable, content-dependent discretization step size.

\section{Dual Refinement Cycle Learning}

\begin{figure}[htbp]
	\centering
	\includegraphics[width=0.47\textwidth]{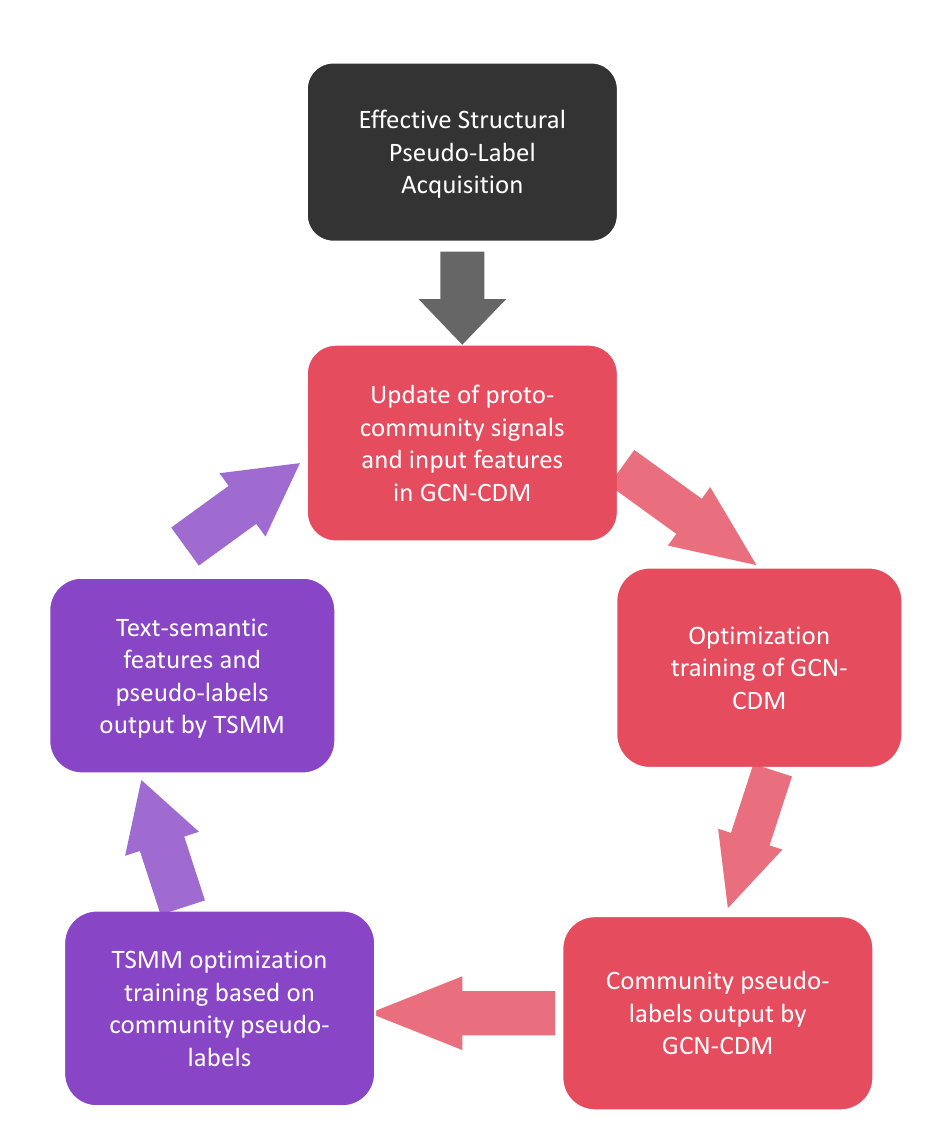}
	\caption{DRCL workflow diagram}
	\label{workflow}
\end{figure}

In real-world networks where ground-truth labels are absent, relying solely on graph structure or node attributes often leads to challenges such as supervision scarcity, noise amplification, and limited interpretability. To address these challenges, we propose a Dual Refinement Cycle Learning (DRCL) framework. It is guided by feature refinement and semantic feedback, and alternately updates two symmetric branches, namely a GCN-based Community Detection Module (GCN-CDM) and a Text Semantic Modeling Module (TSMM). In this way, DRCL constructs a bidirectional closed-loop optimization where the two branches serve simultaneously as each other’s inputs and optimization targets. GCN-CDM and TSMM mutually calibrate each other throughout the iterative cycle, enabling stable convergence and continuous performance improvement under unlabeled conditions. In addition, structural community discovery is incorporated as a warm-start initialization for DRCL, providing the first epoch with a fixed number of labels and interpretable proto-community signals to stabilize the initial training state. The overall workflow is illustrated in Fig.~\ref{workflow}.

\begin{figure}[htbp]
	\centering
	\includegraphics[width=0.75\textwidth]{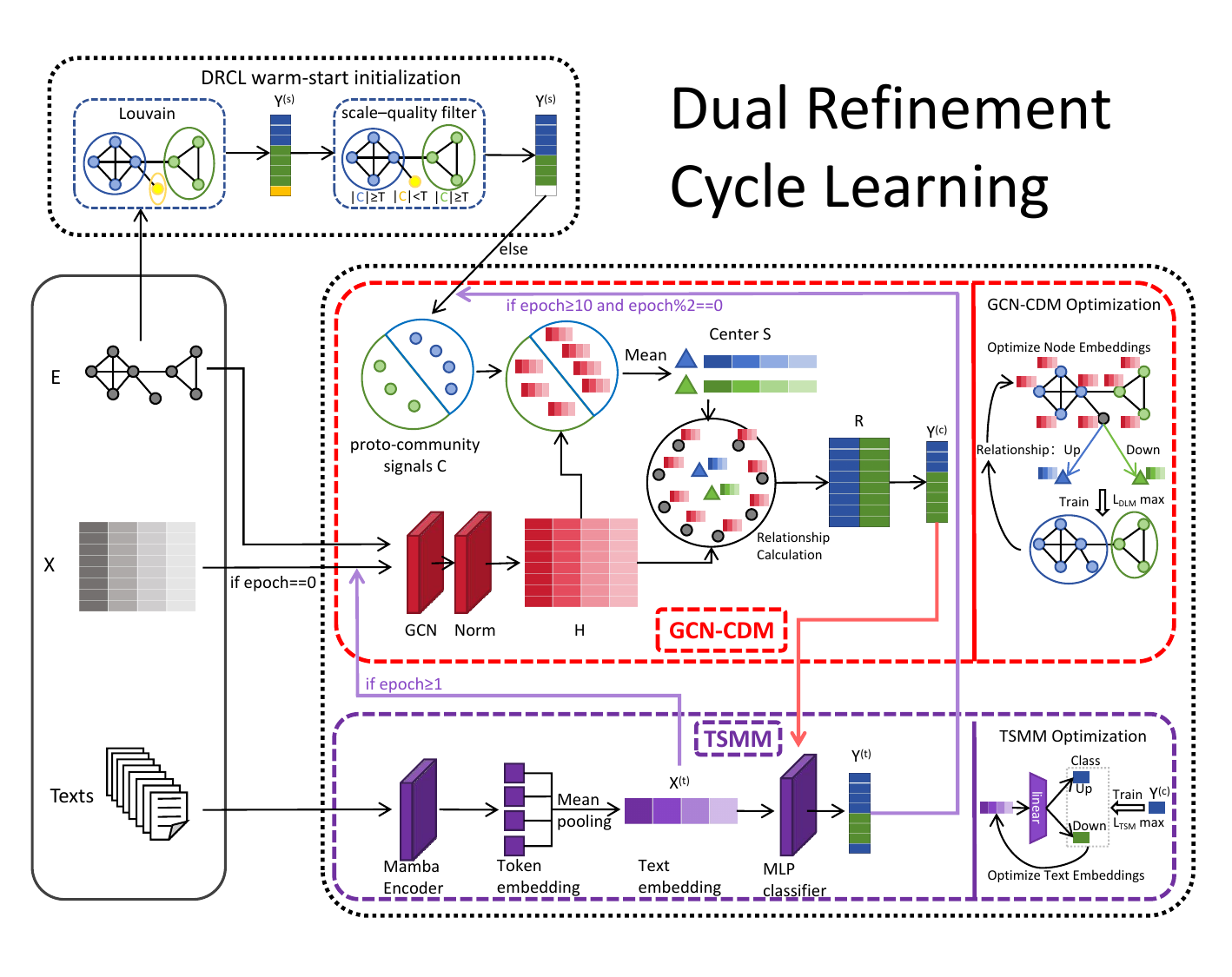}
	\caption{DRCL Framework}
	\label{framework}
\end{figure}

\subsection{Dual Refinement Cycle Learning}

The overall framework of DRCL is illustrated in Fig.~\ref{framework}. It is a closed-loop learning architecture composed of a GCN-CDM and a TSMM.

First, the framework performs a warm-start initialization using the Louvain algorithm together with a scale–quality filtering strategy to determine the number of communities and to generate the initial proto-community signals \(C\) for GCN-CDM. Subsequently, GCN-CDM in DRCL integrates network topology and node attributes, and leverages the proto-community signals to reveal the soft relationships between nodes and communities, thereby producing high-quality community pseudo-labels. Afterward, these community pseudo-labels are used to supervise the Mamba language model within TSMM, enabling it to learn text semantic representations without requiring any ground-truth labels. In turn, the text semantic features and semantic pseudo-labels produced by the trained Mamba model are fed back to GCN-CDM at specific epochs, refining its features and proto-community signals and initiating a new cycle of more accurate community partitioning. Through this cyclic refinement mechanism, the proposed method not only eliminates the dependence of traditional language models on manual annotations, but also enables mutual reinforcement and joint enhancement between topological structure and textual semantics within an unsupervised framework, thereby substantially improving the overall performance of community detection.

Specifically, the community pseudo-labels generated by GCN-CDM, the text semantic features produced by the Mamba model, and the proto-community signals updated from the text semantic pseudo-labels jointly serve as a bridge between GCN-CDM and TSMM, enabling the two modules to engage in closed-loop learning.
\subsection{Proto-Community Signals}

The proto-community signals play a dual role in DRCL, functioning both as the initialization mechanism for the GCN-CDM module and as the bridge that enables TSMM to feed back into GCN-CDM. The proto-community signals are mainly divided into two types, namely those derived from structure-based pseudo-labels and those derived from text-semantic pseudo-labels. DRCL adopts a training strategy that begins with structural stabilization and later transitions into alternating semantic collaboration. During the early epochs, high-quality structure-based pseudo-labels \(Y^{(s)}\) are used as proto-community signals to optimize GCN-CDM, ensuring that the structural information within communities becomes stable. In the later epochs, the framework alternates with the text-semantic labels \(Y^{(t)}\), enabling joint refinement between structural and semantic information. The alternating collaboration ensures that structural consistency and semantic separability constrain and complement each other. This mechanism prevents structural information from suppressing fine-grained semantic distinctions, while also avoiding semantic signals from disrupting topological connectivity.

\paragraph{\textbf{Structure-based Proto-Community Signals}}
High-quality structure-based pseudo-labels \(Y^{(s)}\) are obtained through the structural community discovery algorithm together with a scale–quality filtering strategy, and these pseudo-labels are used as the structure-based proto-community signals \(C^{(s)}\) for GCN-CDM. These high-quality structure-based pseudo-labels do not rely on any prior specification of the number of communities, and they exhibit clear boundaries and strong interpretability in the sense of reflecting high connectivity within communities and low connectivity between them. Building on these properties, they serve not only as the initial pseudo-supervision for both the subsequent GCN-based community detection module and the text semantic modeling process, but also as early-epoch anchors that prevent feature learning from drifting in the absence of ground-truth labels.

In this work, following the framework presented in \cite{RN1}, we obtain the structure-based pseudo-labels \(Y^{(s)}\) as the proto-community signals \(C^{(s)}\) and perform community detection within the GCN-CDM. Specifically, the Louvain algorithm is first applied to partition the network into communities, and each node \(i\) is assigned a structure-based pseudo-label \(y^{(s)}_i\)according to its community assignment. The formulation is given as follows. 
\begin{equation}
	Y^{(s)}=Louvain(A)=[y_1^{(s)},y_2^{(s)},…,y_n^{(s)} ], y_i^{(s)}\in \left \{  1,2,…,l\right \} 
	\label{Louvain}
\end{equation}
where \(l\) denotes the number of communities.

Then, a scale filtering threshold \(T\) is introduced to remove low-quality structure-based pseudo-labels corresponding to communities that are structurally loose or contain too few nodes. After this filtering, node groups with moderate scale and dense internal connectivity are retained and used as the structure-based proto-community signals \(C^{(s)}\):
\begin{equation}
	C^{(s)}=\left \{c_y = \left \{ y_j^{(s)} = y \right \} \mid \left | c_y \right |\ge T  \right \},y\in \left \{ 1,2,...,l \right \} ,j=1,2,...,n 
	\label{C_s}
\end{equation}
where \(\left | c_y \right | \) denotes the size of the current structural cluster. We filter out \(k\) communities set \(C^{(s)}\), thereby retaining the valid structure-based pseudo-labels \(Y^{(s)} \in \left\{1,2,...,k \right\}\).

The threshold \(T\) is jointly determined by the mean \(\mu\) and standard deviation \(\sigma\) of the initial community sizes:
\begin{equation}
	T = \mu + 0.5 \sigma 
	\label{T}
\end{equation}

\paragraph{\textbf{Text-Semantic-based Proto-Community Signals}}
The text-semantic pseudo-labels \(Y^{(t)}\) are generated by the TSMM, as detailed in Section 4.4. The corresponding proto-community signals are defined as follows:
\begin{equation}
	C^{(t)}=\left \{ c_i=\left \{ j\mid y_j^{(t)}=i \right \}  \right \} ,i \in\left \{ 1,2,...,k \right \} ,j=1,2,....n
	\label{C_t}
\end{equation}

To avoid biases caused by long-term dominance of supervision from a single type of proto-community signals, the update rule for the proto-community signals \(C\) in DRCL is defined as follows:
\begin{equation}
	C =\begin{cases}
			C^{(t)}, & \text{if epoch} \ge 10 \ \text{and}\ \text{epoch\%2} = 0, \\
			C^{(s)}, & \text{else}.
		\end{cases}
		\label{C}
\end{equation}

The proto-community signals derived from text-semantic pseudo-labels serve as one of the bridges connecting the two modules, enabling GCN-CDM and TSMM to form a mutually reinforcing learning cycle.

\subsection{GCN-based Community Detection Module}

As the starting point and foundation of DRCL, GCN-CDM undertakes the core responsibility of consolidating structural and feature information. Its goal is to deeply integrate topological structure with attribute features (either numerical features or text semantic features) to produce community partitions that exhibit structural sparsity and attribute similarity. In doing so, it provides clean and reliable pseudo-supervision for the subsequent text semantic modeling. The specific process of this module is shown in algorithm~\ref{alg:GCN-CDM}.

GCN-CDM adopts a GCN as its backbone encoder to integrate topological relations with node features, yielding the structure- and semantics-aware node representations \(H\).
\begin{equation}
	H= Norm(GCN(X^{(I)},A))
	\label{H}
\end{equation}
where \(Norm(\cdot)\) denotes the L2 normalization operation, and \(X^{(I)}\) represents either the initial attribute features of the nodes or the semantic features generated from node texts by the TSMM, as detailed in Eq.~\ref{X_I}.

Subsequently, the proto-community signals \(C\) obtained from Eq.~\ref{C} are used to map the community centers \(S\) in the embedding space.
\begin{equation}
	s_i=\frac{ {\textstyle \sum_{v\in c_i}^{}}h_v }{\left | c_i \right | },i = 1,2,...,k 
	\label{S} 
\end{equation}

Furthermore, the center–node soft relation matrix \(R\) is constructed to reveal the likelihood of each node \(v\) belonging to the different communities.
\begin{equation}
	r_v=softmax(-\delta \cdot (Sh_v) )
	\label{R}
\end{equation}
where \(\delta\) is a temperature hyperparameter that controls the sharpness of the community distribution, smoothing the membership relations and preventing nodes from overly concentrating on a single community.

Finally, more stable community pseudo-labels \(Y^{(c)}\)that are more friendly to boundary nodes are generated and used as pseudo-supervision signals for text semantic modeling.
\begin{equation}
	y_v^{(c)}=argmax(r_v)
	\label{Y_c}
\end{equation}

\begin{algorithm}
	\caption{GCN-CDM}
	\label{alg:GCN-CDM}
	\begin{algorithmic}[1]
		\State \textbf{Input:} Node feature matrix $X^{(I)}$, proto-community signals $C$.
		\State \textbf{Output:} Community pseudo-labels $Y^{(c)}$.
		\For{$l = 1$ to $iter-1$}
		\State {Learn the node representations $H$ in Eq. (\ref{H}).}
		\State {Map community center representations $S$ in Eq. (\ref{S}).}
		\State {Compute relation matrix $R$ and generate community pseudo-labels $Y^{(c)}$ in Eq. (\ref{R}) (\ref{Y_c}).}
		\State {Compute $L_{GCN-CDM}$ and optimize the GCN-CDM in Eq.  (\ref{L_GCN}).}
		\State Calculate the metrics DBI, DI, Q, NMI, ACC, F1 and ARI to evaluate the effectiveness of community partitioning.
		\EndFor
	\end{algorithmic}
\end{algorithm}

\subsection{Text Semantic Modeling Module}

In real-world applications, label scarcity is a prevalent reality, which poses fundamental engineering challenges when Mamba is tasked with unsupervised text classification. Owing to the thinness of training objectives derived solely from self-supervised signals, the model’s optimization direction can easily deviate from the classification goal, leading to insufficient separability of text representations and resulting in a semantic space that struggles to establish clear class boundaries. This engineering bottleneck highlights the urgent need to equip Mamba with high-quality supervisory signals.

Taking the community pseudo-labels produced by GCN-CDM as input, the core objective of TSMM is to incorporate these supervision signals, which originate from network structure and node features, into the semantic modeling of node texts in an effective manner. This enables the learning of high-quality vector representations for node content. Using the community pseudo-labels generated by Eq.~\ref{Y_c} as supervision, the module guides the language model to perform sequence learning tasks and thereby eliminates the need for ground-truth labels during training. The specific process of this module is shown in algorithm~\ref{alg:TSMM}.

TSMM employs Mamba as its backbone encoder \(f_{\theta}(\cdot)\) to perform sequence-level semantic encoding on the text sequence \(t_i\) of node \(i\).
\begin{equation}
	x_i^{(t)}=g(f_\theta (t_i))
	\label{X_t}
\end{equation}
where \(f_\theta(\cdot)\) encodes each token in the text sequence, and \(g(\cdot)\) denotes the mean-pooling layer, which computes the average over all token representations to obtain the text semantic feature \(x_i^{(t)}\) of node \(i\).

Subsequently, a MLP classifier is applied to predict the category of node \(i\):
\begin{equation}
	O_i=Wx_i^{(t)}+b
	\label{O}
\end{equation}
\begin{equation}
	p_i=softmax(O_i)
	\label{P}
\end{equation}
where \(O_i\) denotes the logit vector predicted for node \(i\). \(W\) and \(b\) are the learnable parameters of the MLP classifier, corresponding to its weight matrix and bias term. After applying the softmax function, \(p_i\) gives the predicted probability of node \(i\) for each category. The text-semantic pseudo-label \(y_i^{(t)}\) is then assigned according to the category corresponding to the highest predicted probability in \(p_i\):
\begin{equation}
	y_i^{(t)}=argmax(p_i)
	\label{Y_t}
\end{equation}

The community pseudo-labels \(Y^{(c)}\) generated by GCN-CDM are passed to TSMM as supervisory signals for its text classification training, enabling the trained Mamba model to better capture semantic similarities among texts belonging to the same category. More importantly, this design removes the dependence on manually crafted prior knowledge. The method benefits from two key properties of GCN-CDM. First, the structure-based proto-community signals are derived from a structural community discovery mechanism that can adaptively determine the number of communities. Second, the community pseudo-labels generated by GCN-CDM, through the integration of network topology and node features, exhibit strong discriminative power. Together, these advantages ensure that the entire framework can perform high-quality text classification with the language model without relying on any manual annotations or pre-defined category settings.

Using the trained Mamba model, TSMM extracts the text-semantic features \(X^{(t)}\)and the text-semantic pseudo-labels \(Y^{(t)}\). After completing one round of TSMM training and updating, the resulting semantic features do not serve as the final output. Instead, they flow back into GCN-CDM and replace the input features of the GCN encoder for the next iteration.
\begin{equation}
	X^{(I)}=\begin{cases}
		& X,\text{ if } epoch=0 \\
		& X^{(t)},\text{ else } 
	\end{cases}
	\label{X_I}
\end{equation}

With clearer and more discriminative semantic representations, GCN-CDM can more accurately distinguish confusing categories and boundary nodes when updating community centers and soft memberships, thereby further improving the quality of the community pseudo-labels.

\begin{algorithm}
	\caption{TSMM}
	\label{alg:TSMM}
	\begin{algorithmic}[1]
		\State \textbf{Input:} Text $T$, community pseudo-labels $Y^{(c)}$.
		\State \textbf{Output:} Text-semantic features $X^{(t)}$, text-semantic pseudo-labels $Y^{(t)}$.
		\For{train\_step = 0\% to 100\%}
		\State {Extract the text-semantic features $X^{(t)}$ in Eq. (\ref{X_t}).}
		\State {Predict the text classification probability $P$ in Eq. (\ref{O}) (\ref{P}).}
		\State {Compute $L_{TSMM}$ and update the language model in Eq. (\ref{L_TSMM}).}
		\If{train\_step reaches every 10\%}
			\State Evaluate classification performance on the test set using the ACC metric. 
		\EndIf
	\EndFor
	\State Load the trained language model to extract text-semantic features $X^{(t)}$ and text-semantic pseudo-labels $Y^{(t)}$ in Eq. (\ref{X_t}) (\ref{Y_t}).
	\end{algorithmic}
\end{algorithm}

\subsection{Training Optimization Strategy}

GCN-CDM and TSMM are placed within a tightly coupled joint training paradigm, in which community-distribution optimization and text-semantic optimization act as two primary objectives that drive coordinated learning. The former is constrained by graph structure and community consistency, enabling it to learn community information that exhibits both modularity and separability. The latter centers on semantic discriminability, producing discriminative text embeddings. Together, these components jointly enhance community partitioning by improving structural interpretability and semantic discriminative power.

\paragraph{\textbf{GCN-CDM Training Optimization}}
The optimization core of GCN-CDM lies in refining the relationship between nodes and communities, enabling the model to adjust node embeddings through backpropagation so that they better align with the characteristics of their corresponding communities. The key to this process is to optimize the community partitioning based on the membership probabilities between node embeddings and community centers, thereby ensuring the reasonable assignment of each node to its community. The optimization objective is defined as follows:
\begin{equation}
	L_{GCN-CDM}=-\frac{1}{2M} {\textstyle \sum_{ij} {\textstyle \sum_{k} (a_{ij}-\frac{d_i d_j}{2M} ) r_{ik}r_{jk}} }  
	\label{L_GCN}
\end{equation}
where \(d_i= {\textstyle \sum_{j} a_{ij}}\) denotes the degree of node \(i\), and \(r_{ik}\) represents the membership probability of node \(i\) to community \(k\) in the membership matrix \(R\). By maximizing this objective function, the relationship between nodes and communities is optimized, allowing community-level relational information to guide node representation learning and thereby improving the accuracy and consistency of the community partitioning.

\paragraph{\textbf{TSMM Training Optimization}}
In TSMM, text classification is treated as the core training task, with the goal of learning text semantic vectors that can accurately distinguish between different categories. We split the dataset into training and test sets in a 7:3 ratio, and after every 10\% of the training samples in the training set has been processed, we evaluate on the test set, using the accuracy (ACC) metric to measure the classification performance at the current training stage.

During the training phase, TSMM uses the community pseudo-labels generated by GCN-CDM as supervisory signals to enforce consistency between the predicted distribution and the pseudo-labels. Accordingly, we employ the cross-entropy loss as the optimization objective, which is defined as follows:
\begin{equation}
	L_{TSMM}=-\frac{1}{n}  {\textstyle \sum_{i=1}^{n} \log_{}{p_{i,y_i^{(c)}}} } 
	\label{L_TSMM}
\end{equation}
where \(n\) denotes the total number of nodes, and \(p_{i,y_i^{(c)}}\) represents the probability assigned to the community pseudo-label \(y_i^{(c)}\) in the predicted probability distribution of node \(i\). By maximizing this loss function, the model progressively strengthens its ability to fit the pseudo-labels, thereby acquiring more discriminative and stable semantic feature representations.

Finally, the overall loss objective of the framework is defined as follows:
\begin{equation}
	L=0.001L_{GCN-CDM}+L_{TSMM}
	\label{L}
\end{equation}

The specific process of the proposed framework is shown in algorithm~\ref{alg:DRCL}.

\begin{algorithm}
	\caption{DRCL}
	\label{alg:DRCL}
	\begin{algorithmic}[1]
		\State \textbf{Input:} Text attribute graph $G(V, E, X, T)$.
		\State \textbf{Output:} Community labels.
		\State Perform structural partitioning and obtain the effective structure-based pseudo-labels $Y^{(s)}$ and the proto-community signals $C^{(s)}$ in Eq. (\ref{Louvain}) (\ref{C_s}) (\ref{T}).
		\State Update the input features $X^{(I)}$ and the proto-community signals $C$ of the GCN-CDM in Eq. (\ref{X_I}) and (\ref{C}).
		\For{epoch = 0 to 20}
		\State {Train the GCN-CDM with input node features $X^{(I)}$ and proto-community signals $C$, outputting community pseudo-labels $Y^{(c)}$. //Algorithm~\ref{alg:GCN-CDM}}
		\State {Train the TSMM with input text $T$ and community pseudo-labels $Y^{(c)}$, outputting text-semantic features $X^{(t)}$ and text-semantic pseudo-labels $Y^{(t)}$. //Algorithm~\ref{alg:TSMM}}
		\State {Update the input features $X^{(I)}$ and the proto-community signals $C$ of the GCN-CDM in Eq. (\ref{X_I}) and (\ref{C}).}
		\EndFor
	\end{algorithmic}
\end{algorithm}

\subsection{Equivalence between Mamba Classification and Community Detection}

\noindent \textbf{Theorem 1} {(Convergence Equivalence).} Under the DRCL framework, if the initial proto-community signals in GCN-CDM possess sufficient confidence and the alternating training procedure ensures that both modules reach local convergence in each iteration, then upon the convergence of DRCL, the hard classification labels predicted by the Mamba model in TSMM become equivalent to the community assignments produced by GCN-CDM.

\noindent \textbf{Proof.} In GCN-CDM, the node representations \(H\) are computed according to Eq.~\ref{H}, and the community centers \(S\) are obtained through the mapping defined in Eq.~\ref{S}. The center–node soft relation matrix \(R\) is then calculated using Eq.~\ref{R}. Finally, the high-confidence community pseudo-labels \(Y^{(c)}\) are determined by Eq.~\ref{Y_c}.

TSMM is supervised by the high-confidence pseudo-labels \(Y^{(c)}\) generated by GCN-CDM, enabling Mamba to learn discriminative semantic representations through Eq.~\ref{X_t}. The predicted probability vector \(p_i\) for node \(i\) is then computed using Eq.~\ref{O} and \ref{P}, and the final text-semantic pseudo-labels \(Y^{(t)}\) is determined according to Eq.~\ref{Y_t}.

During each training epoch, Eq.~\ref{L_TSMM} enables the high-confidence pseudo-labels \(Y^{(c)}\) produced by GCN-CDM to supervise the Mamba-based classification in TSMM. After TSMM completes its training, the semantic features learned by Mamba are fed back into the feature learning process of GCN-CDM through Eq.~\ref{X_I}, encouraging the soft relation matrix \(R\) to align with the prediction assignments of Mamba. At convergence, the condition \(y_i^{(c)} = \pi y_i^{(t)}\) holds, where \(\pi\) denotes a permutation mapping over the category indices. This indicates that the text classification performed by Mamba and the community detection produced by GCN-CDM are essentially equivalent. The only possible difference lies in the indexing of category labels, while the actual partitioning of nodes is identical. \(\square\)

\noindent \textbf{Experimental Verification.} This conclusion is empirically supported by Table~\ref{Mamba} in Section 5.5 and Figure~\ref{epoch} in Section 5.6. As the number of refinement cycles increases, the performance of community detection gradually improves and eventually stabilizes, reaching a level comparable to that achieved with medium-scale real annotations. These results further validate the reliability of the stated equivalence relationship.
\subsection{Time Complexity Analysis of DRCL}
The overall computational complexity of DRCL is determined by the combined costs of GCN-CDM and TSMM. For a sparse graph with \(n\) nodes and \(M\) edges, a single GCN propagation layer in GCN-CDM follows Eq.~\ref{GCN}, resulting in a computational cost of 
\(\mathcal{O}\left(M d+n d^{2}\right)\), where \(d\) denotes the dimensionality of the node features. Since real-world graphs are typically sparse and satisfy \(M \propto n\), this term can be regarded as scaling linearly .

In the semantic branch, TSMM employs Mamba to encode the node texts. Unlike the \(\mathcal{O}{(L^2)}\) attention complexity of Transformers, Mamba performs sequence modeling of length \(L\) in linear time. Consequently, the encoding cost for each node is \(\mathcal{O}{(Ld_s)}\), and for the entire graph it becomes \(\mathcal{O}{(nLd_s)}\), where \(d_s\) denotes the dimensionality of the hidden states.

DRCL performs 20 refinement cycles during training, with each cycle consisting of one structural update and one semantic update. Therefore, the overall computational complexity is \(\mathcal{O}{(20(Md+nL^2+nLd_s))}\). Since both core modules exhibit linear scalability, DRCL maintains high efficiency in large-scale graph–text scenarios and demonstrates strong practical scalability.

\section{Experimentation}
The experiments were conducted on a computer with an Intel i9 processor, an NVIDIA GeForce RTX2080Ti GPU, and the Ubuntu 16.04 operating system, using a Python 3.10 environment for programming and computation.

\subsection{Datasets and Evaluation Metrics}
To comprehensively evaluate the model’s generalization capability across networks of varying scale, structure, and textual semantics, this study adopts six publicly available textual datasets, covering three major categories: citation networks, encyclopedic networks, and social media networks. 

\textbf{Citation Networks.} This category includes Cora\cite{RN65}, Citeseer\cite{RN66}, Pubmed\cite{RN67}, and Arxiv\_2023\cite{RN51}. In these datasets, nodes represent papers and edges denote citation relationships. The original node features are bag-of-words or TF–IDF vectors, and node labels correspond to research topics. The textual content associated with each node consists of the paper title and abstract. The four datasets increase in scale sequentially, and although they exhibit relatively clear community structures, they also contain varying degrees of class imbalance.

\textbf{Encyclopedic Network.} WikiCS \cite{RN68}. In this dataset, nodes correspond to Wikipedia pages, and edges represent hyperlinks between pages. The original node features are TF–IDF or Doc2Vec vectors, the node labels denote computer science subdisciplines, and the textual content consists of the titles and main bodies of the Wikipedia articles.

\textbf{Social Media Network.} Instagram \cite{RN69}. In this dataset, nodes represent users, and edges denote the following relationships between users. The original node features consist of extracted vectors from user post images and content. Node labels correspond to commercial user label information, and the textual content includes the users' personal profiles.

To ensure the completeness and comparability of the evaluation, this paper selects six widely used metrics for comprehensive assessment: NMI, ACC, F1, ARI, DBI and Q. Among these, DBI and Q measure the quality of community partitioning from the perspectives of internal clustering cohesion and graph structure consistency, respectively. NMI, ACC, F1, and ARI are external consistency metrics commonly used in both community detection and language text classification tasks, reflecting the degree of alignment between the obtained categories and the ground-truth categories. 

\subsection{Baseline Methods and Parameter Settings}
To comprehensively evaluate the performance of the proposed method, we selected two representative community detection baselines: deep graph representation learning methods and language model text encoding methods. The former directly learns node representations that fuse structural and attribute information on the graph, while the latter uses pre-trained language models to extract text attribute vectors and then combines them with unsupervised clustering to obtain community partitions.

\textbf{Deep Learning-based Methods.} (a) In CommDGI, node embeddings are learned by employing a mutual-information objective between local node-level representations and a global summary vector of the whole graph. (b) DGCluster casts modularity maximization as a differentiable loss term, which is optimized together with a GNN encoder so that representation learning and clustering are trained in a single end-to-end pipeline. (c) DCGL builds a pseudo-siamese network with two branches that separately encode structural information and attribute information; the two views are then aligned by cross-view contrastive learning to strengthen the final embeddings. (d) MAGI regards the modularity matrix as an anchor in a contrastive learning framework, enabling the model to capture high-order structural similarity without explicitly constructing augmented graphs. (e) MGCN introduces multi-hop graph convolution modules that adaptively aggregate signals from higher-order neighborhoods, leading to richer and more informative node representations. (f) CPGCL jointly optimizes node embeddings and soft community assignment variables, and continuously reselects contrastive sample pairs during training so as to alleviate the problem of false negatives.

\textbf{Language Model-based Methods.} (a) BERT introduced the bidirectional Transformer encoding and masked language model pretraining paradigm, making bidirectional context modeling mainstream. By enhancing sentence-level objectives to strengthen inter-sentence relational representations, it significantly improves general semantic extraction capabilities. (b) SciBERT pre-trains on large-scale scientific corpora and constructs a domain-specific vocabulary to reduce token fragmentation. By learning with the same pretraining objective as BERT on different statistical distributions, it demonstrates the benefits of corpus transfer and vocabulary adaptation for the representation of term-dense texts. (c) Qwen3-embedding-0.6B employs a lightweight encoder and is trained using contrastive learning and retrieval optimization objectives. It combines hard negative sample mining and multi-granularity pooling, enhancing fine-grained semantic differentiation while ensuring inference efficiency. Its multilingual alignment and quantization-friendly design make it suitable for low-cost deployment. (d) Jina-embeddings-v2-base-code uses cross-domain contrastive training and instruction-based embedding alignment, balancing semantic spaces for both natural language and program semantics. It adopts code-aware tokenization and multi-view positive sampling strategies, improving consistency and robustness across different expression forms. (e) Mamba2 replaces self-attention with a selective state-space model for linear-time long sequence modeling. It introduces gating and selective routing to dynamically allocate computation, improving the stability of long-context processing and throughput, demonstrating better scalability and hardware friendliness in sequence encoding. 

The proposed framework is composed of GCN-CDM and TSMM and is trained for 20 epochs. GCN-CDM uses a single-layer graph convolution network to learn node representations, with the hyperparameter \(\delta \) fixed at 30. Model parameters are optimized with Adam for 300 update steps, using a learning rate of 0.001 and a weight decay of 0.005. For TSMM, we adopt the AdamW optimizer and choose dataset-specific learning rates: 5e-5 for the Cora dataset, 5e-6 for Citeseer, WikiCS, and Instagram datasets, and 5e-7 for Pubmed and Arxiv\_2023 datasets. A cosine learning-rate schedule with a warmup ratio of 0.001 is applied to stabilize the initial phase and to enable smooth annealing. In all experiments, the number of communities is set equal to the number detected by our framework on each dataset.

For deep learning–based baselines, we closely follow the hyperparameter settings reported in the original works to ensure fair comparison. Specifically, CommDGI is trained with a learning rate of 0.001 for 500 iterations; DGCluster and DCGL both use a learning rate of 0.001 with 300 iterations; MAGI adopts a learning rate of 0.0005 with 400 iterations; MGCN uses a learning rate of 0.003 with 700 iterations; and CPGCL is trained with a learning rate of 0.0007 for 600 iterations.

The language model extracts node text vectors from the textual content using its pre-trained model. Subsequently, the obtained embeddings undergo L2 normalization, followed by K-means clustering to obtain the community partitioning results.

\subsection{Experimental Results}
The experimental results of the proposed algorithm compared to other methods are shown in Table~\ref{result}. Overall, the proposed method achieves optimal or highly competitive results across most datasets and evaluation metrics. Compared to existing deep learning methods, our approach not only integrates network structure and node features, but more importantly, it introduces text semantic information and continuously refines it through the dual refinement cycle. This results in significant advantages in the community detection task. These findings validate the effectiveness of the mutual enhancement between text semantics and topology, and demonstrate that the dual refinement strategy can more comprehensively capture all the information embedded in text-attributed graphs.	
%\clearpage
\begin{center}
	\small
	\renewcommand{\arraystretch}{1.4} % 缩小行间距，默认为1
	\setlength{\tabcolsep}{1pt}
\begin{longtable}{ccccccccc}
	\caption{Performance comparison of the different algorithms. “OM” denotes an out-of-memory failure, “N/A” means the algorithm runs for more than five days, and “NAN” indicates that a NAN error occurred during execution. Symbols marked with \(\uparrow\) represent metrics for which larger values are better, while \(\downarrow\) indicates metrics where smaller values are preferred. {\ul Underlined values} indicate the second-best results, while \textbf{boldface values} denote the best results.}
	\\
	\toprule
	&                      &                                               & DBI($\downarrow$)          & Q($\uparrow$)                  & NMI($\uparrow$)          & ACC($\uparrow$)          & F1($\uparrow$)           & ARI($\uparrow$)          \\
	\midrule
	\endfirsthead
	
	\multicolumn{9}{l}{\small\slshape Continued from previous page} \\
	\toprule
	&                      &                                               & DBI($\downarrow$)          & Q($\uparrow$)                  & NMI($\uparrow$)          & ACC($\uparrow$)          & F1($\uparrow$)           & ARI($\uparrow$)          \\
	\midrule
	\endhead
	
	\midrule
	\multicolumn{9}{r}{\small\slshape Continued on next page} \\
	\endfoot
	
	\endlastfoot

	\multirow{12}{*}{Cora}       
	& Deep Learning-based  & CommDGI                                       & 1.564707          & 0.695584                & 0.512530          & 0.653988          & 0.642800          & 0.426486          \\
	&                      & DCGL                                          & 1.725745          & 0.177538                & 0.157732          & 0.376292          & 0.337400          & 0.062756          \\
	&                      & MAGI                                          & 1.730136          & 0.717382                & 0.590037          & 0.750369          & 0.771500          & 0.532026          \\
	&                      & MGCN                                          & 0.483558          & 0.185911                & 0.211655          & 0.374446          & 0.340185          & 0.077713          \\
	&                      & DGCluster                                     & 0.947829          & \textbf{0.751538}       & 0.462999          & 0.282127          & 0.235300          & 0.154965          \\
	&                      & CPGCL                                         & 2.700266          & 0.506380                & 0.250061          & 0.427253          & 0.435109          & 0.166023          \\
	\cline{2-9}
	& Language Model-based & Bert                                          & 3.188392          & 0.064343                & 0.025875          & 0.207533          & 0.226006          & 0.011548          \\
	&                      & SciBert                                       & 2.804513          & 0.098046                & 0.047201          & 0.234490          & 0.261928          & 0.018954          \\
	&                      & Qwen3-embedding-0.6B                          & 4.380674          & 0.490378                & 0.423962          & 0.562408          & 0.598936          & 0.331315          \\
	&                      & Jina-embeddings-v2-base-code                  & 3.738668          & 0.335491                & 0.246019          & 0.433161          & 0.479717          & 0.180700          \\
	&                      & Mamba2                                        & 1.617823          & 0.137311                & 0.083274          & 0.289882          & 0.287935          & 0.046961          \\
%	\cline{2-9}
	& DRCL                 & \multicolumn{1}{l}{Without Semantic Feedback} & {\ul 0.202715}    & 0.723385                & {\ul 0.597160}    & {\ul 0.760709}    & {\ul 0.772630}    & \textbf{0.585694} \\
	&                      & Semantic Feedback                             & \textbf{0.162303} & {\ul 0.725636} & \textbf{0.603979} & \textbf{0.768095} & \textbf{0.782111} & {\ul 0.584886}    \\
	\pagebreak
%	\midrule
	\multirow{12}{*}{Citeseer}
	& Deep Learning-based  & CommDGI                                       & 1.496301          & {\ul 0.743162}          & 0.424532          & 0.622724          & 0.540200          & 0.412054          \\
	&                      & DCGL                                          & 1.226520          & 0.636256                & 0.422776          & 0.635593          & 0.526090          & 0.411193          \\
	&                      & MAGI                                          & 1.134440          & 0.719462                & 0.396710          & 0.529504          & 0.528437          & 0.368415          \\
	&                      & MGCN                                          & 0.492561          & 0.019468                & 0.019742          & 0.220025          & 0.183757          & 0.008753          \\
	&                      & DGCluster                                     & 1.143336          & \textbf{0.825254}       & 0.339573          & 0.171061          & 0.168100          & 0.096286          \\
	&                      & CPGCL                                         & 2.061462          & 0.450728                & 0.321074          & 0.494036          & 0.457865          & 0.272126          \\
	\cline{2-9}
	& Language Model-based & Bert                                          & 3.392044          & 0.054351                & 0.042277          & 0.173572          & 0.200392          & 0.023763          \\
	&                      & SciBert                                       & 3.199334          & 0.042773                & 0.035795          & 0.170433          & 0.194570          & 0.017668          \\
	&                      & Qwen3-embedding-0.6B                          & 4.084227          & 0.523474                & 0.379415          & 0.488073          & 0.486447          & 0.327463          \\
	&                      & Jina-embeddings-v2-base-code                  & 3.658114          & 0.445085                & 0.297799          & 0.356874          & 0.422179          & 0.210465          \\
	&                      & Mamba2                                        & 1.889747          & 0.155648                & 0.103959          & 0.241055          & 0.268749          & 0.063274          \\
	\cline{2-9}
	& DRCL                 & Without Semantic Feedback                     & \textbf{0.319680} & 0.728750                & {\ul 0.426539}    & {\ul 0.675141}    & \textbf{0.639632} & {\ul 0.445914}    \\
	&                      & Semantic Feedback                             & {\ul 0.324130}    & 0.723163                & \textbf{0.432592} & \textbf{0.677966} & {\ul 0.638612}    & \textbf{0.450242} \\
	\midrule
	\multirow{12}{*}{Wikics}
	& Deep Learning-based  & CommDGI                                       & 2.039129          & 0.584923                & 0.468502          & {\ul 0.615930}    & 0.527600          & 0.445070          \\
	&                      & DCGL                                          & 1.648628          & 0.418684                & 0.465857          & 0.560550          & 0.539228          & 0.359747          \\
	&                      & MAGI                                          & 1.386865          & 0.572777                & {\ul 0.511922}    & \textbf{0.621913} & \textbf{0.622517} & 0.451531          \\
	&                      & MGCN                                          & 0.575866          & 0.155776                & 0.081101          & 0.250662          & 0.167529          & 0.038082          \\
	&                      & DGCluster                                     & 1.435376          & 0.569237                & 0.491026          & 0.443466          & 0.314200          & 0.351912          \\
	&                      & CPGCL                                         & 2.922549          & 0.563907                & 0.324086          & 0.465601          & 0.472168          & 0.211654          \\
	\cline{2-9}
	& Language Model-based & Bert                                          & 3.011186          & 0.148604                & 0.170385          & 0.264251          & 0.300232          & 0.089550          \\
	&                      & SciBert                                       & 3.610762          & 0.067505                & 0.068227          & 0.162550          & 0.196512          & 0.033114          \\
	&                      & Qwen3-embedding-0.6B                          & 4.659534          & 0.394563                & 0.402490          & 0.439450          & 0.485627          & 0.267341          \\
	&                      & Jina-embeddings-v2-base-code                  & 3.347417          & 0.234279                & 0.291763          & 0.356294          & 0.392462          & 0.165429          \\
	&                      & Mamba2                                        & 1.772533          & 0.147586                & 0.147172          & 0.216819          & 0.253320          & 0.073256          \\
	\cline{2-9}
	& DRCL                 & Without Semantic Feedback                     & \textbf{0.384167} & \textbf{0.629184}       & 0.510580          & 0.614819          & 0.596175          & {\ul 0.487943}    \\
	&                      & Semantic Feedback                             & {\ul 0.415174}    & {\ul 0.628361}          & \textbf{0.512972} & 0.615588          & {\ul 0.603117}    & \textbf{0.502681} \\
	\midrule
	\multirow{6}{*}{Pubmed}
	& Deep Learning-based  & CommDGI                                       & 1.177054          & \textbf{0.625854}       & 0.286372          & 0.607800          & 0.568393          & 0.257209          \\
	&                      & DCGL                                          & N/A               & N/A                     & N/A               & N/A               & N/A               & N/A               \\
	&                      & MAGI                                          & 1.916181          & {\ul 0.613154}          & 0.212856          & 0.544606          & 0.534488          & 0.166876          \\
	&                      & MGCN                                          & 0.673912          & 0.164186                & 0.031439          & 0.418573          & 0.340158          & 0.002792          \\
	&                      & DGCluster                                     & 1.570793          & 0.576274                & 0.223114          & 0.124461          & 0.119900          & 0.053200          \\
	&                      & CPGCL                                         & OOM               & OOM                     & OOM               & OOM               & OOM               & OOM               \\
%	\cline{2-9}
	\pagebreak
	\multirow{7}{*}{Pubmed}
	& Language Model-based & Bert                                          & 3.090685          & 0.223496                & 0.087697          & 0.451134          & 0.447498          & 0.085156          \\
	&                      & SciBert                                       & 2.891342          & 0.168153                & 0.064890          & 0.391185          & 0.410246          & 0.056515          \\
	&                      & Qwen3-embedding-0.6B                          & 3.945258          & 0.508963                & 0.266861          & 0.568849          & 0.592811          & 0.225795          \\
	&                      & Jina-embeddings-v2-base-code                  & 4.387049          & 0.258082                & 0.084988          & 0.393772          & 0.421597          & 0.060858          \\
	&                      & Mamba2                                        & 1.517157          & 0.333971                & 0.178322          & 0.490186          & 0.485505          & 0.117872          \\
	\cline{2-9}
	& DRCL                 & Without Semantic Feedback                     & \textbf{0.185938} & 0.567338                & {\ul 0.302482}    & \textbf{0.680276} & {\ul 0.679070}    & {\ul 0.291010}    \\
	&                      & Semantic Feedback                             & {\ul 0.227610}    & 0.563854                & \textbf{0.318398} & {\ul 0.678957}    & \textbf{0.679474} & \textbf{0.295725} \\
	\midrule
	\multirow{12}{*}{Arxiv\_2023} 
	& Deep Learning-based  & CommDGI                                       & 2.390256          & 0.520749                & 0.263115          & 0.475670          & 0.266376          & 0.225748          \\
	&                      & DCGL                                          & N/A               & N/A                     & N/A               & N/A               & N/A               & N/A               \\
	&                      & MAGI                                          & OOM               & OOM                     & OOM               & OOM               & OOM               & OOM               \\
	&                      & MGCN                                          & OOM               & OOM                     & OOM               & OOM               & OOM               & OOM               \\
	&                      & DGCluster                                     & 1.356785          & 0.349374                & 0.340109          & 0.129421          & 0.103500          & 0.031319          \\
	&                      & CPGCL                                         & OOM               & OOM                     & OOM               & OOM               & OOM               & OOM               \\
	\cline{2-9}
	& Language Model-based & Bert                                          & 3.499630          & 0.069387                & 0.136825          & 0.133707          & 0.018551          & 0.044692          \\
	&                      & SciBert                                       & 3.021439          & 0.082192                & 0.145898          & 0.135569          & 0.033313          & 0.041178          \\
	&                      & Qwen3-embedding-0.6B                          & 4.238073          & 0.458584                & \textbf{0.413564} & 0.274016          & 0.021668          & 0.148360          \\
	&                      & Jina-embeddings-v2-base-code                  & 3.805632          & 0.311242                & 0.315625          & 0.249145          & 0.007684          & 0.131812          \\
	&                      & Mamba2                                        & 2.041672          & 0.136892                & 0.219890          & 0.193861          & 0.222588          & 0.080071          \\
	\cline{2-9}
	& DRCL                 & Without Semantic Feedback                     & \textbf{0.487308} & \textbf{0.532605}       & 0.330519          & {\ul 0.502922}    & \textbf{0.448242} & {\ul 0.330416}    \\
	&                      & Semantic Feedback                             & {\ul 0.567635}    & {\ul 0.530952}          & {\ul 0.350075}    & \textbf{0.516278} & {\ul 0.417535}    & \textbf{0.402997} \\
%	\pagebreak
	\midrule
	\multirow{12}{*}{Instagram}   
	& Deep Learning-based  & CommDGI                                       & 2.940017          & {\ul 0.512543}          & {\ul 0.013811}    & {\ul 0.574654}    & 0.393795          & \textbf{0.050665} \\
	&                      & DCGL                                          & N/A               & N/A                     & N/A               & N/A               & N/A               & N/A               \\
	&                      & MAGI                                          & 1.696688          & 0.260475                & 0.001952          & 0.536555          & \textbf{0.568920} & 0.004920          \\
	&                      & MGCN                                          & 0.617742          & 0.216320                & 0.005420          & \textbf{0.580034} & 0.368433          & {\ul 0.010698}    \\
	&                      & DGCluster                                     & 1.411628          & \textbf{0.512750}       & \textbf{0.020007} & 0.050798          & 0.048500          & 0.000909          \\
	&                      & CPGCL                                         & NAN               & NAN                     & NAN               & NAN               & NAN               & NAN               \\
	\cline{2-9}
	& Language Model-based & Bert                                          & 2.951994          & 0.030028                & 0.000064          & 0.511862          & {\ul 0.548604}    & 0.000579          \\
	&                      & SciBert                                       & 3.050893          & 0.055049                & 0.002867          & 0.198342          & 0.244920          & 0.000318          \\
	&                      & Qwen3-embedding-0.6B                          & 5.174326          & 0.160535                & 0.003314          & 0.156539          & 0.221664          & 0.001917          \\
	&                      & Jina-embeddings-v2-base-code                  & 2.974532          & 0.040905                & 0.005445          & 0.215716          & 0.256619          & 0.006907          \\
	&                      & Mamba2                                        & 1.452264          & 0.053819                & 0.005765          & 0.232648          & 0.272931          & 0.004912          \\
	\cline{2-9}
	& DRCL                 & Without Semantic Feedback                     & \textbf{0.607371} & 0.488642                & 0.005677          & 0.433372          & 0.353160          & 0.009234          \\
	&                      & Semantic Feedback                             & {\ul 0.575601}    & 0.470651                & 0.006813          & 0.441838          & 0.367109          & 0.010666          \\ \bottomrule
	\label{result}
\end{longtable}
\end{center}

\subsection{MCDM Ranking}
To assess the algorithms from a holistic perspective, we adopt the TOPSIS-based multi-criteria decision-making (MCDM) framework. Since the six evaluation indicators used in our experiments differ in type and exhibit non-negligible correlations, we employ the CRITIC method to automatically derive objective weights, thereby avoiding subjective bias from manual assignment. The procedure is as follows. First, the raw metric values are preprocessed by handling special symbols (e.g., OM, N/A, NAN) and then normalized to remove inconsistencies in scale and optimization direction. Next, CRITIC is applied to compute the weight of each metric by jointly considering its discriminative power and information redundancy. Finally, the TOPSIS scheme is used to compute the closeness of each algorithm to the positive and negative ideal solutions, and the algorithms are ranked according to these proximity scores, yielding a unified and objective performance ordering under the multi-dimensional evaluation system. The detailed results are reported in Table~\ref{MCDM}. 

Experimental evidence on multiple real-world datasets shows that this evaluation pipeline can reliably capture the overall performance gaps among competing methods. The approach proposed in this study achieves the top rank on most datasets, demonstrating its clear advantage in terms of aggregated performance across multiple metrics.

\begin{table*}[htbp]
	\centering
	\caption{MCDM Ranking}
	\begin{adjustbox}{width=1\textwidth}
		\begin{tabular}{ccccccccccccccc} 
			\toprule 
			&              & CommDGI         & DCGL   & MAGI   & MGCN   & DGCluster & CPGCL  & Bert   & SciBert & Qwen3-embedding-0.6B & Jina-embeddings-v2-base-code & Mamba2 & Without Semantic Feedback & Semantic Feedback \\
			\midrule
			Cora        & Topsis Score & 0.7447          & 0.3799 & 0.7878 & 0.4852 & 0.5385    & 0.4183 & 0.1594 & 0.2135  & 0.4284               & 0.3176                       & 0.3551 & 0.9810                    & \textbf{0.9857}   \\
			& Rank         & 4               & 9      & 3      & 6      & 5         & 8      & 13     & 12      & 7                    & 11                           & 10     & 2                         & \textbf{1}        \\
			\midrule
			Citeseer    & Topsis Score & 0.7910          & 0.8086 & 0.7963 & 0.4375 & 0.5295    & 0.5893 & 0.1171 & 0.1423  & 0.4556               & 0.3836                       & 0.3567 & 0.9573                    & \textbf{0.9590}   \\
			& Rank         & 5               & 3      & 4      & 9      & 7         & 6      & 13     & 12      & 8                    & 10                           & 11     & 2                         & \textbf{1}        \\
			\midrule
			Wikics      & Topsis Score & 0.7494          & 0.7454 & 0.8520 & 0.4552 & 0.6767    & 0.5444 & 0.2910 & 0.1482  & 0.4225               & 0.3772                       & 0.3901 & 0.9747                    & \textbf{0.9819}   \\
			& Rank         & 4               & 5      & 3      & 8      & 6         & 7      & 12     & 13      & 9                    & 11                           & 10     & 2                         & \textbf{1}        \\
			\midrule
			Pubmed      & Topsis Score & 0.8581          & 0.0000 & 0.6958 & 0.3406 & 0.4832    & 0.0000 & 0.3277 & 0.2729  & 0.6356               & 0.2714                       & 0.5170 & 0.9354                    & \textbf{0.9370}   \\
			& Rank         & 3               & 12     & 4      & 8      & 7         & 12     & 9      & 10      & 5                    & 11                           & 6      & 2                         & \textbf{1}        \\
			\midrule
			Arxiv\_2023 & Topsis Score & 0.6541          & 0.0000 & 0.0000 & 0.0000 & 0.4087    & 0.0000 & 0.0754 & 0.1225  & 0.4501               & 0.3402                       & 0.3204 & 0.8752                    & \textbf{0.9151}   \\
			& Rank         & 3               & 10     & 10     & 10     & 5         & 10     & 9      & 8       & 4                    & 6                            & 7      & 2                         & \textbf{1}        \\
			\midrule
			Instagram   & Topsis Score & \textbf{0.7715} & 0.0000 & 0.5518 & 0.5631 & 0.5143    & 0.0000 & 0.4206 & 0.2428  & 0.2052               & 0.2719                       & 0.3474 & 0.6337                    & 0.6452            \\
			& Rank         & \textbf{1}      & 12     & 5      & 4      & 6         & 12     & 7      & 10      & 11                   & 9                            & 8      & 3                         & 2                
			\\ \bottomrule
		\end{tabular}
	\end{adjustbox}	                  
	\label{MCDM}
\end{table*}

\subsection{Analysis of Mamba Classification Performance Driven by Community Detection}

To investigate the impact of community detection results as supervisory signals on language model training, this study designs a comparative experimental setup. The community pseudo-labels generated by the GCN-CDM are used as supervisory signals for the Mamba model, alongside real labeled data at different proportions (10\%, 30\%, 50\%), and the model's performance is systematically evaluated.

The experimental results, as shown in Table~\ref{Mamba}, confirm the effectiveness of community detection as a supervisory signal in the training of the Mamba model. It not only achieves performance comparable to or even superior to that of training with moderate proportions of real-label supervision, but also enables the Mamba model to break free from the constraints of real labels. This provides a reliable technical path for large-scale applications in real-world networks, with significant theoretical value and practical significance.
\begin{table*}[htbp]
	\centering
	\caption{Experimental results of Mamba with different supervision training methods}
	\begin{adjustbox}{width=0.85\textwidth}
		\begin{tabular}{ccccc} 
			\toprule
			\multicolumn{1}{l}{} & \multicolumn{1}{l}{}                 & ACC($\uparrow$)          & F1($\uparrow$)           & ARI($\uparrow$)          \\
			\midrule
			Cora                 & Community Pseudo-label Supervision   & \textbf{0.768095} & \textbf{0.782111} & \textbf{0.584886} \\
			& 10\% Real Label Supervision Training & 0.318685          & 0.127400          & 0.069188          \\
			& 30\% Real Label Supervision Training & 0.266617          & 0.142171          & 0.031145          \\
			& 50\% Real Label Supervision Training & 0.317947          & 0.108567          & 0.062094          \\
			\midrule
			Citeseer             & Community Pseudo-label Supervision   & \textbf{0.677966} & \textbf{0.638612} & \textbf{0.450242} \\
			& 10\% Real Label Supervision Training & 0.242624          & 0.111425          & 0.074428          \\
			& 30\% Real Label Supervision Training & 0.278406          & 0.117702          & 0.087771          \\
			& 50\% Real Label Supervision Training & 0.279033          & 0.087257          & 0.104749          \\
			\midrule
			Wikics               & Community Pseudo-label Supervision   & \textbf{0.615588} & \textbf{0.603117} & 0.502681          \\
			& 10\% Real Label Supervision Training & 0.249039          & 0.069054          & 0.103249          \\
			& 30\% Real Label Supervision Training & 0.496111          & 0.010597          & 0.354123          \\
			& 50\% Real Label Supervision Training & 0.603531          & 0.059397          & \textbf{0.507738} \\
			\midrule
			Pubmed               & Community Pseudo-label Supervision   & 0.678957          & \textbf{0.679474} & 0.295725          \\
			& 10\% Real Label Supervision Training & 0.718162          & 0.292692          & 0.425924          \\
			& 30\% Real Label Supervision Training & 0.871735          & 0.228584          & 0.735780          \\
			& 50\% Real Label Supervision Training & \textbf{0.877923} & 0.206319          & \textbf{0.759428} \\
			\midrule
			Arixv\_2023          & Community Pseudo-label Supervision   & \textbf{0.516278} & \textbf{0.417535} & 0.402997          \\
			& 10\% Real Label Supervision Training & 0.338478          & 0.020023          & 0.241136          \\
			& 30\% Real Label Supervision Training & 0.507684          & 0.011970          & 0.422863          \\
			& 50\% Real Label Supervision Training & 0.500134          & 0.009026          & \textbf{0.531469} \\
			\midrule
			Instagram            & Community Pseudo-label Supervision   & \textbf{0.441838} & \textbf{0.367109} & \textbf{0.010666} \\
			& 10\% Real Label Supervision Training & 0.110768          & 0.047447          & 0.002642          \\
			& 30\% Real Label Supervision Training & 0.105212          & 0.066849          & 0.002381          \\
			& 50\% Real Label Supervision Training & 0.129729          & 0.059000          & 0.003884   \\
			\bottomrule
		\end{tabular}
	\end{adjustbox}	                  
	\label{Mamba}
\end{table*}

\subsection{Analysis of Iterative Rounds in Dual Refinement Cycle Learning}
\begin{figure}[htbp]
	\centering 
	\begin{minipage}[b]{0.45\linewidth}
		\centering
		\includegraphics[width=\linewidth]{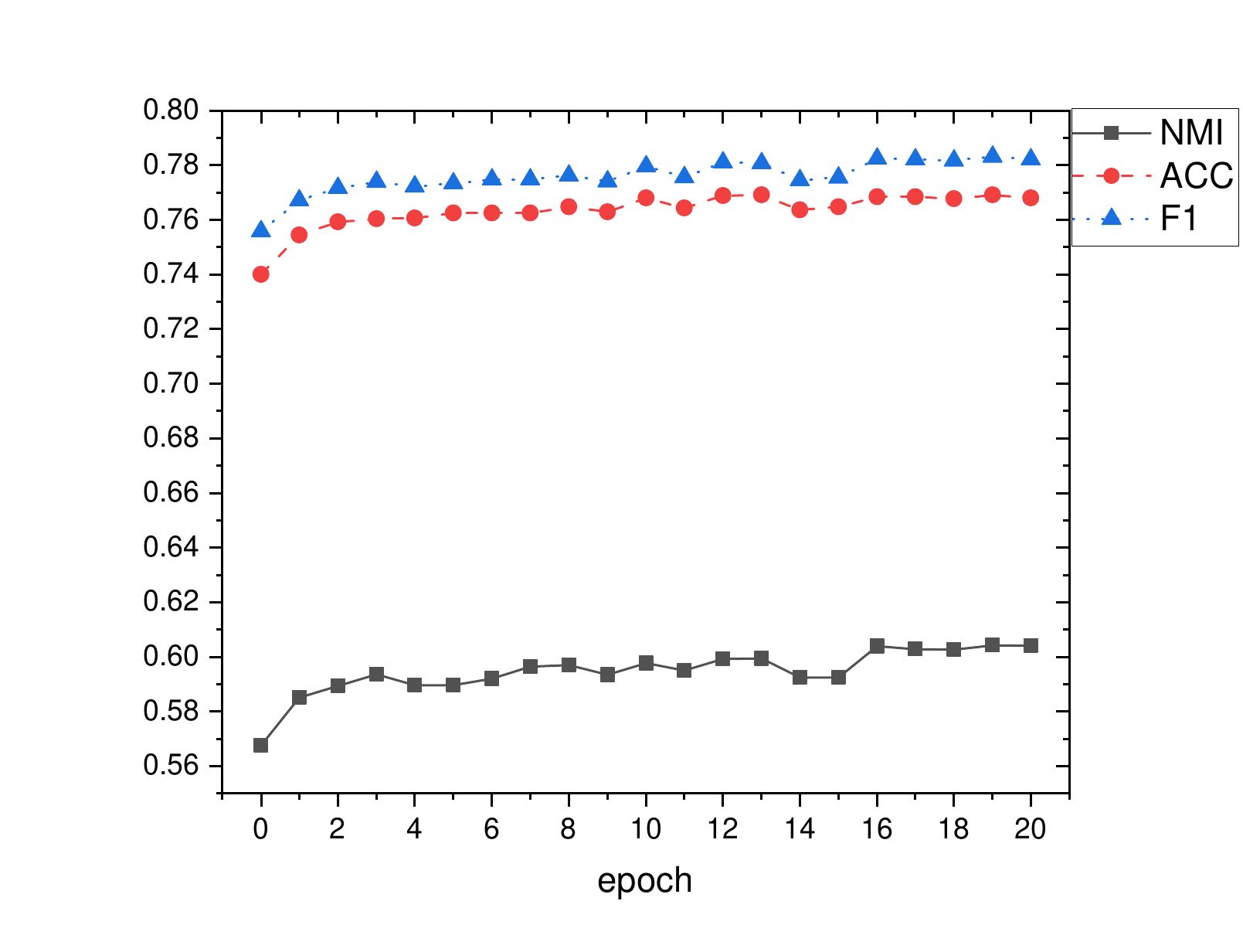}
		\subcaption{Cora} 
	\end{minipage}
	\begin{minipage}[b]{0.45\linewidth}
		\centering
		\includegraphics[width=\linewidth]{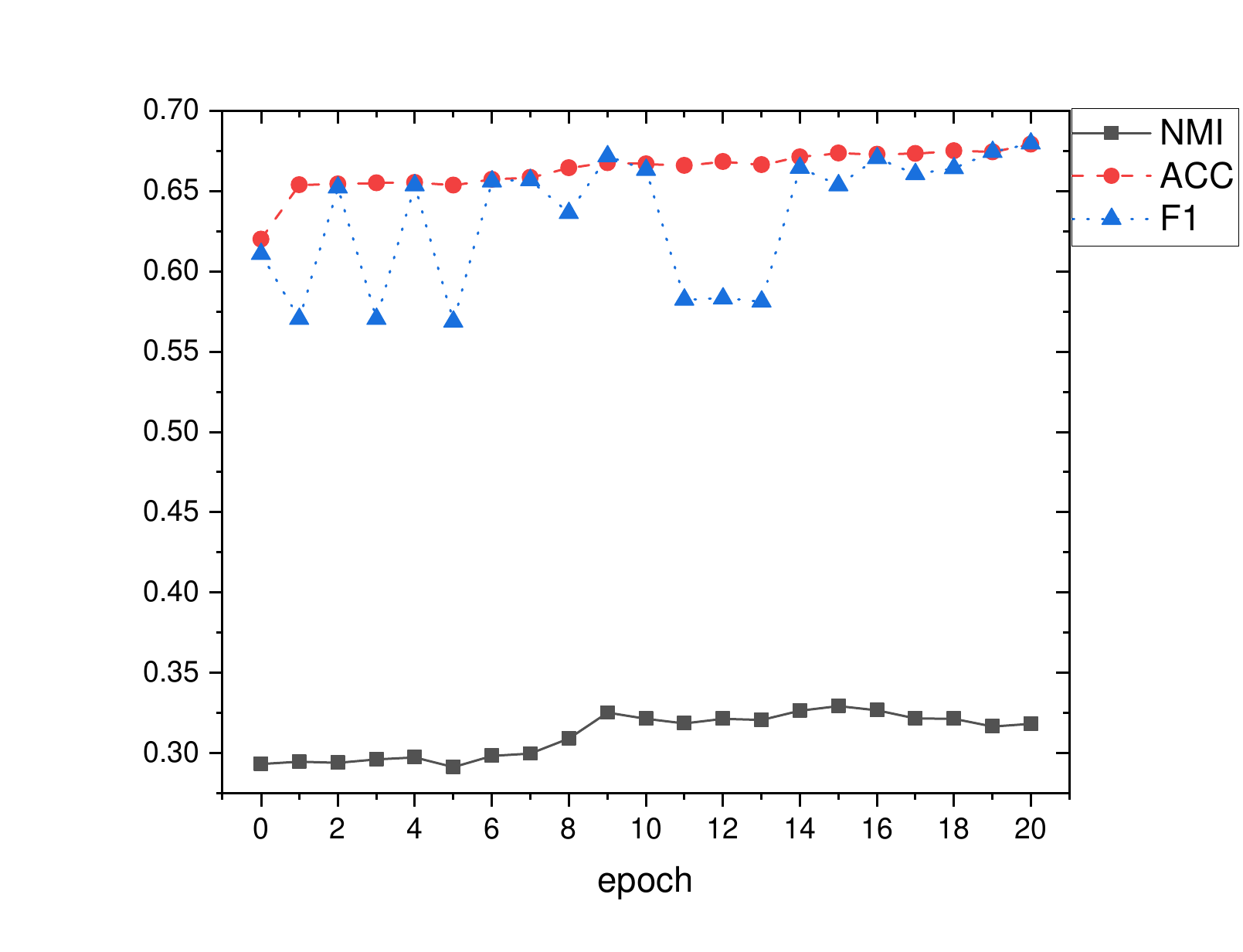}
		\subcaption{Pubmed} 
	\end{minipage}	
	
	\caption{Dynamic optimization process of Dual Refinement Cycle Learning} 
	\label{epoch}
\end{figure}

To further investigate the dynamic optimization process of the Dual Refinement Cycle Learning, this study selected two datasets, Cora and PubMed, and conducted a tracking analysis of the model over 20 complete training epochs based on three key metrics: NMI, ACC, and F1. The experimental results, as shown in Fig.~\ref{epoch}, indicate that in the first 10 training epochs, the model primarily relies on high-quality structural pseudo-labels as pre-community signals. During this phase, the structural information is effectively reinforced, and the performance metrics show a steady improvement. Subsequently, text semantic pseudo-labels and high-quality structural pseudo-labels alternate in updating the pre-community signals, further promoting the optimization of the classification boundary. All metrics stabilize between epochs 18 and 20. 

\subsection{Parameter Sensitivity Analysis of the GCN-CDM-Side Optimization Objective}
To examine the appropriateness of the weighting assigned to the GCN-CDM-side optimization objective \(L_{GCN-CDM}\) in Eq.~\ref{L}, this section reformulates the overall loss objective in a parameterized form: 
\begin{equation}
	L = \lambda L_{GCN-CDM} + L_{TSMM}
\end{equation}

We then conduct a parameter sensitivity analysis with respect to the coefficient \(\lambda\). Experiments are performed on the Cora and Citeseer datasets, where the network architecture and all other hyperparameters are fixed. Only the value of \(\lambda\) is varied within {1.0, 0.1, 0.01, 0.001}, and the model performance is evaluated using the metrics Q, NMI, ACC, and F1. 

\begin{figure}[htbp]
	\centering 
	\begin{minipage}[b]{0.45\linewidth}
		\centering
		\includegraphics[width=\linewidth]{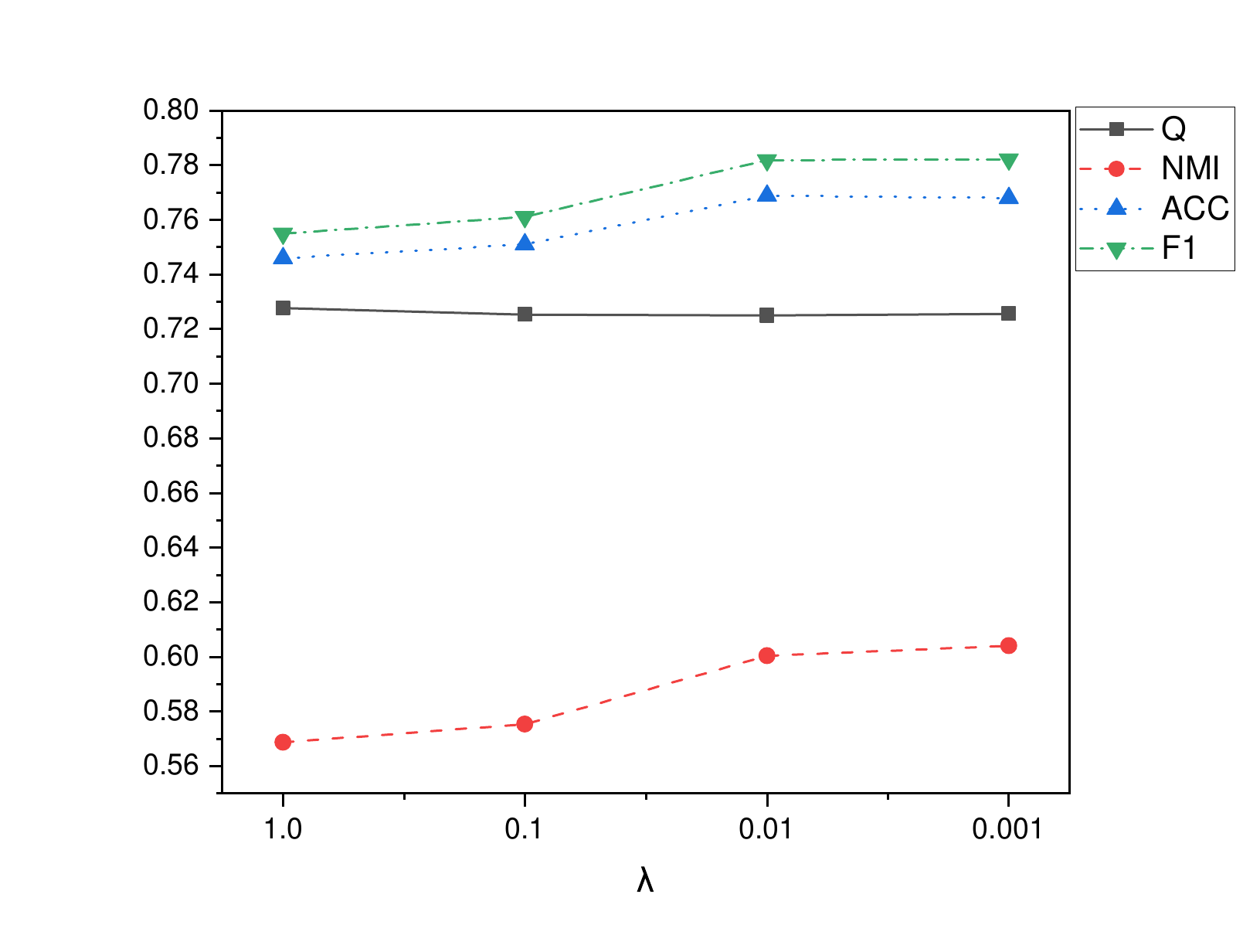}
		\subcaption{Cora} 
	\end{minipage}
	\begin{minipage}[b]{0.45\linewidth}
		\centering
		\includegraphics[width=\linewidth]{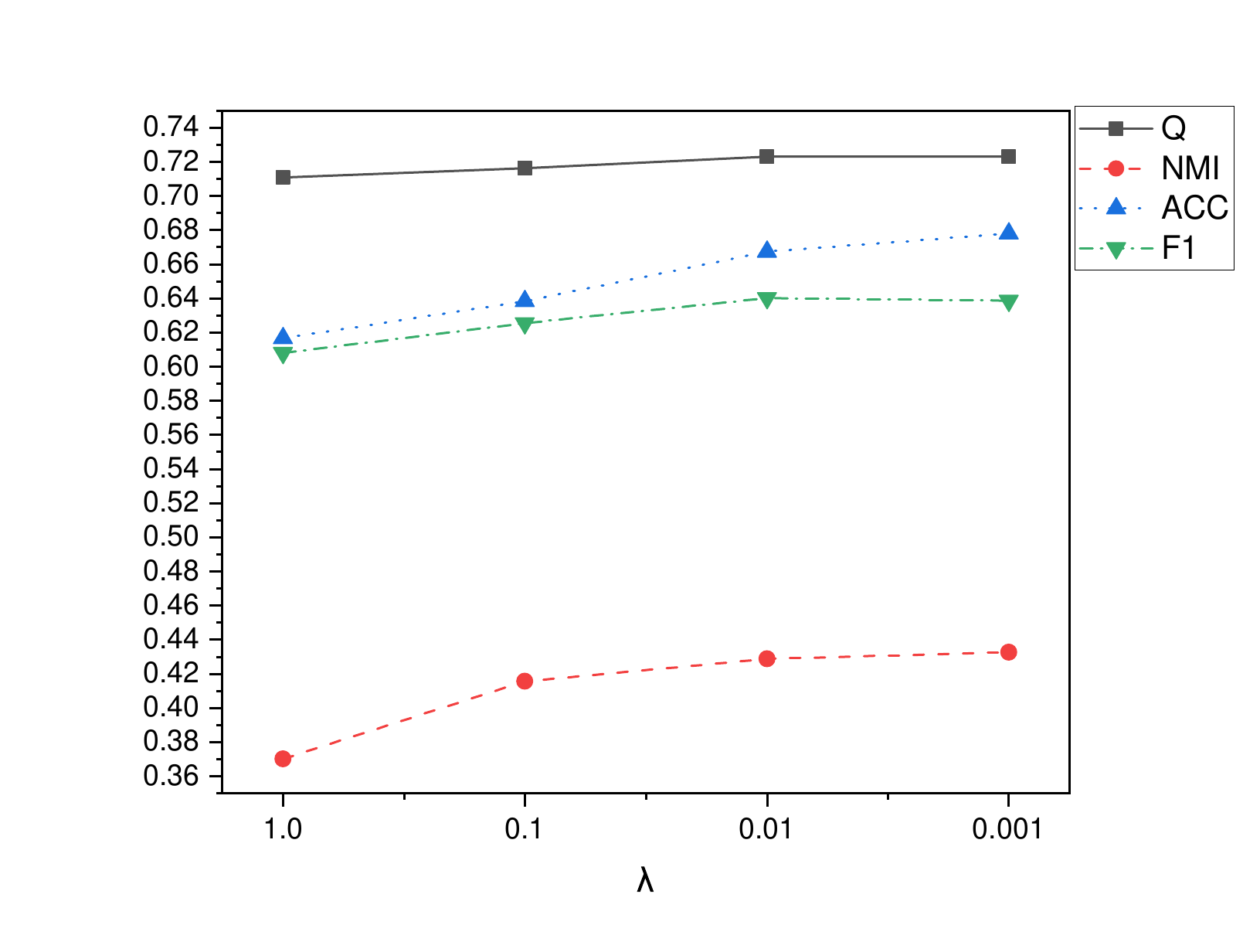}
		\subcaption{Citeseer} 
	\end{minipage}	
	
	\caption{Parameter sensitivity analysis of the loss weight \(\lambda\) for \(L_{GCN-CDM}\) on the Cora and Citeseer datasets.} 
	\label{para}
\end{figure}

The results are shown in Fig.~\ref{para}. As \(\lambda\) decreases, the overall model performance exhibits an upward trend. In particular, when \(\lambda\) is set to 0.001, the model achieves the best or near-best performance on both datasets. Therefore, we fix the loss weight of \(L_{GCN-CDM}\) to 0.001 in the remaining experiments.

\subsection{Ablation Studies}

\subsubsection{Ablation Analysis of TSMM}
To verify the key role of TSMM in the overall framework, this subsection designs a comparative experiment: while keeping other settings unchanged, community detection is performed using only GCN-CDM, with TSMM and its semantic feedback mechanism removed.
 
The experimental results, as shown in Table~\ref{TSMM}, indicate that the performance of the model decreases after removing TSMM, fully demonstrating the necessity of the TSMM module. The TSMM module not only enhances the representation capability of nodes through text semantic features, but more importantly, it provides continuous optimization power for community detection through the cyclic refinement mechanism. 

\begin{table*}[htbp]
	\centering
	\caption{Ablation experiment results of TSMM}
	\begin{adjustbox}{width=0.75\textwidth}
		\begin{tabular}{cccccccc} 
			\toprule
			&              & DBI($\downarrow$)          & Q($\uparrow$)            & NMI($\uparrow$)          & ACC($\uparrow$)          & F1($\uparrow$)           & ARI($\uparrow$)          \\
			\midrule
			Cora        & Without TSLL & 0.518072          & \textbf{0.729730} & 0.567524          & 0.740030          & 0.755850          & 0.534452          \\
			& With TSLL    & \textbf{0.162303} & 0.725636          & \textbf{0.603979} & \textbf{0.768095} & \textbf{0.782111} & \textbf{0.584886} \\
			\midrule
			Citeseer    & Without TSLL & 0.432800          & \textbf{0.777627} & \textbf{0.434697} & 0.612053          & 0.580937          & 0.432320          \\
			& With TSLL    & \textbf{0.324130} & 0.723163          & 0.432592          & \textbf{0.677966} & \textbf{0.638612} & \textbf{0.450242} \\
			\midrule
			Wikics      & Without TSLL & 0.710040          & \textbf{0.628633} & 0.503028          & 0.604307          & 0.585585          & 0.479505          \\
			& With TSLL    & \textbf{0.415174} & 0.628361          & \textbf{0.512972} & \textbf{0.615588} & \textbf{0.603117} & \textbf{0.502681} \\
			\midrule
			Pubmed      & Without TSLL & 0.581284          & \textbf{0.650354} & 0.293325          & 0.619973          & 0.610863          & 0.266285          \\
			& With TSLL    & \textbf{0.227610} & 0.563854          & \textbf{0.318398} & \textbf{0.678957} & \textbf{0.679474} & \textbf{0.295725} \\
			\midrule
			Arxiv\_2023 & Without TSLL & 1.095555          & 0.460959          & 0.204377          & 0.351162          & 0.246508          & 0.198183          \\
			& With TSLL    & \textbf{0.567635} & \textbf{0.530952} & \textbf{0.350075} & \textbf{0.516278} & \textbf{0.417535} & \textbf{0.402997} \\
			\midrule
			Instagram   & Without TSLL & 0.684267          & \textbf{0.513729} & \textbf{0.007804} & 0.416703          & 0.333898          & \textbf{0.014957} \\
			& With TSLL    & \textbf{0.575601} & 0.470651          & 0.006813          & \textbf{0.441838} & \textbf{0.367109} & 0.010666       \\
			\bottomrule  
			
		\end{tabular}
	\end{adjustbox}	                  
	\label{TSMM}
\end{table*}

\subsubsection{Ablation Analysis of GCN-CDM}
To systematically evaluate the contribution of the GCN-CDM in the overall framework, this experiment removes GCN-CDM after the full training cycle and performs k-means clustering using only the text features extracted by TSMM, in order to assess the performance of pure semantic features in the community detection task.

The experimental results, as shown in Table~\ref{GCN-CDM}, indicate that the model's performance decreases after removing GCN-CDM, clearly revealing the important value of GCN-CDM in integrating structural information. TSMM enhances the discriminative ability for boundary nodes through text semantic understanding, while the GCN-CDM module ensures the structural rationality of community division through topological structure information. The collaborative effect of the two modules in the DRCL achieves high-performance unsupervised community detection. 

\begin{table*}[htbp]
	\centering
	\caption{Ablation experiment results of GCN-CDM}
	\begin{adjustbox}{width=0.75\textwidth}
		\begin{tabular}{cccccccc} 
		\toprule
		&                 & DBI($\downarrow$)          & Q($\uparrow$)            & NMI($\uparrow$)          & ACC($\uparrow$)          & F1($\uparrow$)           & ARI($\uparrow$)          \\
		\midrule
		Cora        & Without GCN-CDM & 0.200989          & 0.723327          & 0.598183          & 0.761448          & 0.773862          & \textbf{0.585462} \\
		& With GCN-CDM    & \textbf{0.162303} & \textbf{0.725636} & \textbf{0.603979} & \textbf{0.768095} & \textbf{0.782111} & 0.584886          \\
		\midrule
		Citeseer    & Without GCN-CDM & 1.440825          & 0.55883           & 0.370513          & 0.553045          & 0.515063          & 0.366366          \\
		& With GCN-CDM    & \textbf{0.32413}  & \textbf{0.723163} & \textbf{0.432592} & \textbf{0.677966} & \textbf{0.638612} & \textbf{0.450242} \\
		\midrule
		Wikics      & Without GCN-CDM & 0.915159          & 0.596449          & 0.496284          & 0.60106           & 0.575458          & 0.476269          \\
		& With GCN-CDM    & \textbf{0.415174} & \textbf{0.628361} & \textbf{0.512972} & \textbf{0.615588} & \textbf{0.603117} & \textbf{0.502681} \\
		\midrule
		Pubmed      & Without GCN-CDM & 0.956297          & 0.519531          & 0.313856          & 0.66856           & 0.606519          & 0.293834          \\
		& With GCN-CDM    & \textbf{0.22761}  & \textbf{0.563854} & \textbf{0.318398} & \textbf{0.678957} & \textbf{0.679474} & \textbf{0.295725} \\
		\midrule
		Arxiv\_2023 & Without GCN-CDM & 1.666361          & 0.277522          & 0.311058          & 0.349496          & 0.319309          & 0.247112          \\
		& With GCN-CDM    & \textbf{0.567635} & \textbf{0.530952} & \textbf{0.350075} & \textbf{0.516278} & \textbf{0.417535} & \textbf{0.402997} \\
		\midrule
		Instagram   & Without GCN-CDM & 1.65179           & 0.212626          & 0.004092          & 0.182997          & 0.232552          & 0.000654          \\
		& With GCN-CDM    & \textbf{0.575601} & \textbf{0.470651} & \textbf{0.006813} & \textbf{0.441838} & \textbf{0.367109} & \textbf{0.010666}
		\\ \bottomrule
		\end{tabular}
	\end{adjustbox}	                  
	\label{GCN-CDM}
\end{table*}

\section{Conclusion}
This study addresses the critical challenge of unsupervised text classification and community detection on Text-Attributed Graphs, specifically under conditions where ground-truth labels and the number of categories are unknown. Existing methodologies often fail to adequately leverage the complementary nature of structural topology and deep textual semantics, leading to issues of supervision scarcity and insufficient semantic utilization. To effectively bridge the methodological gap between graph structure analysis and sequence-based semantic modeling, we introduce the novel Dual Refinement Cycle Learning (DRCL) framework.

Experiments on multiple text-attributed graph datasets show that DRCL consistently outperforms deep community detection baselines and enables Mamba-based text classification to reach or surpass the performance of supervised models trained on medium-scale labeled data. DRCL thus provides an effective and general paradigm for jointly leveraging graph structure and textual semantics in fully unsupervised scenarios, laying the groundwork for future research on weakly or fully unsupervised learning in text-attributed graphs.

\bibliographystyle{elsarticle-num}
\bibliography{elsarticle}



%\begin{thebibliography}{00}
%
%%% For authoryear reference style
%%% \bibitem[Author(year)]{label}
%%% Text of bibliographic item
%
%\bibitem[Lamport(1994)]{lamport94}
%  \textit{\LaTeX: a document preparation system},
%  Addison Wesley, Massachusetts,
%\end{thebibliography}
\end{document}